\documentclass[final,journal,letterpaper]{IEEEtran}
 \usepackage[hidelinks]{hyperref}
 \usepackage{varioref} %smart page, figure, table, and equation referencing
 \usepackage{wrapfig} %wrap figures/tables in text (i.e., Di Vinci style)
 \usepackage{threeparttable} %tables with footnotes
 \usepackage{dcolumn} %decimal-aligned tabular math columns
  \newcolumntype{d}{D{.}{.}{-1}}
 \usepackage{nomencl} %nomenclature generation via makeindex
  \makeglossary
 \usepackage{subfigmat} % matrices of similar subfigures, aka small mulitples
 \usepackage{fancyvrb} %extended verbatim environments
  \fvset{fontsize=\footnotesize,xleftmargin=2em}
 \usepackage{lettrine} %dropped capital letter at beginning of paragraph

 \usepackage{epsfig}
 \usepackage{epstopdf}
 \usepackage{subfigure}
 \usepackage{latexsym}
 \usepackage{amsmath}
 \usepackage{amsfonts}
 \usepackage{amssymb}
 \usepackage{mathrsfs}
 \usepackage{cases}
 \usepackage{array}
 \usepackage{color}
 \usepackage{soul}
 \setulcolor{red}
 \sethlcolor{yellow}
 \renewcommand\hl[1]{#1} %  commont out this for highlighted version

 \usepackage{url}
 \usepackage[noadjust]{cite}
 \usepackage{graphicx}
 \usepackage{psfrag}
 \usepackage{color}
   \usepackage{multirow}
 \usepackage{ragged2e}
 \usepackage{graphicx}  

\newfont{\Bb}{msbm10 scaled\magstep1}

\begin{document}

\title{Adaptive Power System Emergency Control using Deep Reinforcement Learning}

\author{Qiuhua Huang, \textit{Member, IEEE}, Renke Huang, \textit{Member, IEEE}, Weituo Hao, Jie Tan,  Rui Fan, \textit{Member, IEEE}, and Zhenyu Huang, \textit{Fellow, IEEE}%
\thanks{\hrule width 2in}
\thanks{The Pacific Northwest National Laboratory (PNNL) is operated by Battelle for the U.S. Department of Energy under Contract DE-AC05-76RL01830. This work was supported by Deep Science laboratory directed research and development at PNNL. Corresponding author: Renke Huang (renke.huang@pnnl.gov)}
\thanks{Q. Huang, R. Huang, R. Fan and Z. Huang are with Pacific Northwest National Laboratory, Richland, WA 99354, USA (e-mail: \{qiuhua.huang, renke.huang,  rui.fan, zhenyu.huang\}@pnnl.gov).}
\thanks{W. Hao is with Department of Electrical and Computer Engineering, Duke University, Durham, NC, 27708, USA  (e-mail: weituo.hao@duke.edu).}
\thanks{Jie Tan is with Google Brain, Google Inc, Mountain View, CA, 94043 USA (e-mail: jietan@google.com).}}
%\thanks{R. Diao is with Global Energy Interconnection Research Institute North America, San Jose, CA 95134, USA (e-mail: Ruisheng.Diao@geirina.net).}}%

\maketitle

\begin{abstract}
Power system emergency control is generally regarded as the last safety net for grid security and resiliency. Existing emergency control schemes are usually designed off-line based on either the conceived ``worst'' case scenario or a few typical operation scenarios. These schemes are facing significant adaptiveness and robustness issues as increasing uncertainties and variations occur in modern electrical grids. To address these challenges, this paper developed novel adaptive emergency control schemes using deep reinforcement learning (DRL)  by leveraging the high-dimensional feature extraction and non-linear generalization capabilities of DRL for complex power systems. Furthermore, an open-source platform named Reinforcement Learning for Grid Control (RLGC) has been designed for the first time to assist the development and benchmarking of DRL algorithms for power system control. Details of the platform and DRL-based emergency control schemes for generator dynamic braking and under-voltage load shedding are presented. \hl{Robustness of the developed DRL method to  different simulation scenarios, model parameter  uncertainty  and noise in the observations is investigated.} Extensive case studies performed in both the two-area, four-machine system and the IEEE 39-bus system have demonstrated excellent performance and robustness of the proposed schemes.
\end{abstract}
 

\section{Introduction}\label{sec:Intro}
\IEEEPARstart{R}{ELIABLE}  and resilient electricity is vital to the economy and national security of all countries. Preventive control measures have been widely employed to ensure adequate security margins against some conceived (e.g. N-1) contingencies. However, several large blackouts still occurred in the US, Europe, India and Brazil in the last two decades \cite{Bo2015, Makarov2005, Ferc2012}. It has been well recognized that emergency control is imperative in real-time operation to minimize the occurrence and impact of power outages or wide-spread blackouts. Conventional emergency control actions include generation redispatch or tripping, load shedding, controlled system separation (or islanding), and dynamic braking \cite{Kundur2000}.

Some of these actions are automatically triggered by control or protection systems, while others are armed by system operators. Ideally, these emergency control actions should be adaptive to real-time system operation conditions.  However, existing control and protection systems for emergency controls are usually based on fixed settings that are mostly determined off-line based on some typical scenarios, and they are operated in a ``set-and-forget'' mode. Emergency controls used by system operators in control rooms today are predefined through off-line studies based on a few forecasted system conditions and conceived contingency scenarios. In addition, it heavily relies on system operators to choose suitable control actions by matching the current system situation with the nearest system conditions defined in emergency control look-up tables, as well as determining when and how to apply them. These processes are time consuming and often overwhelming for system operators. For example, during the 11-minute time span of the 2011 Southwest blackout event in US, system operators lacked sufficient time to understand the causes and take effective corrective actions \cite{Ferc2012}.

Current research into solutions to the emergency control problem can be categorized into three directions: 1) security-constrained alternating current optimal power flow (SC-ACOPF) \cite{Misra2017}; 2) optimal control \cite{Li2017}; and 3) conventional machine learning,  such as decision tree \cite{Genc2010} and conventional reinforcement learning (RL) \cite{Ernst2004, Glavic2017}.

Mathematically, power system emergency control is a problem of dynamic, sequential decision-making under-uncertainty. When being applied to solve this problem, SC-ACOPF is inherently limited by its static formulation of the problem and poor scalability. Optimal control-based methods are generally difficult to scale to handle large-scale power systems and a large number of control actions, and  are not adaptive to system uncertainties. RL methods can solve sequential decision-making problems in real time \cite{Sutton1998}. \hl{The last two  decades have seen increasing efforts to apply conventional RL methods, such as Q-learning and fitted Q-iteration {\cite{Sutton1998}},
in various decision-making and control problems in power systems; these range from demand response  {\cite{RL_DR_review_2019}}, 
energy management, and automatic generation control to transient stability and emergency control {\cite{Glavic2017},\cite{RL_emergency_control},\cite{RL_load_shedding}}. 
Due to scalability issues, applications of conventional RL methods are mainly focusing on problems with low-dimensional state and action spaces.} In addition, their performance is heavily dependent on the quality of handcrafted features \cite{Mnih2015}. Thus, they are not suitable for large, complex problems, such as emergency control for large-scale power systems.

In the past few years, significant progress has been made in solving challenging problems in games \cite{Mnih2015,Silver2017}, robotics \cite{Amarjyoti2017}, etc. using deep reinforcement learning (DRL), which is a combination of deep learning technologies and RL. Unlike conventional RL, by replacing the hand-crafted feature mapping and extraction (such as Q-table) with deep learning technologies, DRL enables automatic high-dimensional feature extraction and end-to-end learning through stochastic gradient descent. In addition, the high-dimensional feature representation capability of deep learning technologies and the development of scalable learning algorithms such as Deep Q-Network (DQN) \cite{Mnih2015} and Proximal Policy Optimization (PPO) \cite{Schulman2017} significantly improve the scalability of DRL, making it suitable for solving large-scale control problems, such as Dota II \cite{OpenAI}. \hl{These advantages of DRL were recognized by some researchers and leveraged in several different applications in power systems in the past few years. In {\cite{DRL_MG_management}}, authors utilized a DRL method to optimize the operation of storage devices in a microgrid considering both future electricity consumption and  photovoltaic (PV) output uncertainties. A DRL approach was applied to solve the  problem of jointly determining the energy bid submitted to the wholesale market and the energy price charged in the retail  market for a load serving entity in {\cite{DRL_Pricing_2019}}. DRL was applied to develop a dynamic load shedding scheme for short-term voltage control in {\cite{DRL_LS_CSG}}. In {\cite{DRL_Gen_trip}}, authors applied DRL to determine generation unit tripping under emergency circumstances.} 
%% what are the gaps??
In light of these, we proposed to develop adaptive and robust power system emergency control schemes using DRL.

One main challenge faced by both power system and RL research communities is reproducing and benchmarking existing work and accurately judging the improvements offered by novel RL methods \cite{henderson2018deep}.  Open platforms and/or tools, such as OpenAI Gym \cite{Brockman2016} and ELF \cite{Tian2017} have been proven to be significantly beneficial for developing and comparing RL algorithms in games and robotics. On the other hand, to the best knowledge of the authors, all previous research efforts in application of RL for power system control \cite{Glavic2017} were based on simulation environments and RL algorithms that were not publicly available. Lack of an open platform for developing, training and benchmarking DRL algorithms for power system control not only becomes a roadblock for power system researchers and engineers to work on applying DRL in power systems, but also prevents many researchers with machine learning and control backgrounds from easily applying their DRL algorithms in power system control. To fill this gap and address the reproducibility issue, an open platform named RLGC \cite{RLGC} for developing, training and benchmarking DRL algorithms for power system control has been developed in this paper. To the best knowledge of the authors, this is the first of this kind platform in the power system area. It is extensible, lightweight and flexible. 

The main contributions of this paper include: 1)  novel application of DRL algorithms for power system emergency controls, including generator dynamic braking and under-voltage load shedding (UVLS); 2) development of the first open-source platform for developing and benchmarking DRL algorithms for power system control; 3) \hl{detailed investigation into several important aspects of DRL algorithms for grid control, including adapting generic problem formulations and the DQN algorithm to detailed, specific emergency control designs, robustness to different simulation scenarios, model parameter uncertainty, and noise in the input (observation), with direct comparisons with a conventional Q-learning and an optimal control methods.}
%study of several important aspects of DRL algorithms for grid emergency control, including robustness to noise in the input (observation) and generalization capability.

The rest of the paper is organized as follows: an overview of DRL and grid emergency control is presented in Section \ref{sec:Overview}.  Section \ref{sec:Platform} details the open platform for developing and benchmarking DRL algorithms for grid control;  Section \ref{sec:Algorithm} discusses development details of two DRL-based grid emergency control schemes; Test cases and results are shown in Section \ref{sec:Results}; discussions on several key aspects of DRL applications for grid emergency control are presented in Section \ref{sec:discussion}; and conclusions and future work are provided in Section \ref{sec:Conc}.

%%%%%%%%%%%%%%%%%%%%%%%%%%%%%%%%%%%%%%%%%%%%%%%%%%%%%%%%%%%%%%%%%%%%%%%%%%%%%%%%%%%%%%%%%%%%
\section{Overview of Deep Reinforcement Learning and Grid Emergency Control}\label{sec:Overview}

\subsection{Reinforcement Learning}
In RL, the agent learns to make optimal decisions by interacting with the environment through exploration and exploitation \cite{Sutton1998}. \hl{The environment is modeled as a (partially observable) Markov decision process (MDP), defined by:}

\begin{itemize}
    \item \hl{a state space $\mathcal{S}$ that could be continuous or discrete}; 
    \item \hl{an action space $\mathcal{A}$ that could be continuous or discrete};
    \item \hl{an environment transition function  $\mathcal{P} :\mathcal{S} \times \mathcal{A} \longrightarrow \mathcal{S}$};
    \item \hl{a reward function $\mathcal{R} :\mathcal{S} \times \mathcal{A} \longrightarrow \mathcal{R}$};
    \item \hl{a discount factor $\gamma \in [0,1]$}.
\end{itemize}

\hl{In this setting, at each time step  $t$, the agent can observe the state  $s_t \in \mathcal{S} $ and receive reward signals $r_t \in \mathcal{R} $ from the environment. At the same time, the agent can select an action $a_t\in \mathcal{A} $ to change the environment.} The goal is to apply the optimal action given the current state so that the agent can accumulate most rewards over time, which are generally defined as discounted future return $R_t$.
\vspace{-1mm}
\begin{equation}
R_t = \sum^T_{t'=t} \gamma ^{t' - t} r_{t'}
\end{equation} \label{eq:futurereturn}%
where $T$ means the time step when the interaction with the system ends. To evaluate the result of the action based on current state, the action-value function also known as Q function,  is proposed as $Q(s,a)$. We define the optimal Q-value of the state-action pair $(s,a)$ as $Q^*(s,a)$, which represents the maximum discounted future return after taking action $a$ at state $s$.
The Q function is updated by the iteration algorithm in the Bellman equation, defined by  \cite{Sutton1998}
\vspace{-2mm} 
\begin{equation}
Q_{t+1}(s,a) = E\left[ r + \gamma max_{a'}Q_t(s',a')|(s,a)   \right] 
\vspace{-4mm} 
\end{equation} \label{eq:Bellamn}

The iteration will converge to the optimal solution $Q^*(s,a)$ as $t \rightarrow \infty$ if the state signals have the Markov property \cite{Sutton1998}.

Q-Learning \cite{Sutton1998} is a value-based RL algorithm which finds the optimal action-selection policy using

\vspace{-4mm} 
\begin{equation}
Q(s_t,a_t) \leftarrow  Q(s_t,a_t)  + \eta\left[ r_{t+1} + \gamma max_{a}Q(s_{t+1},a) - Q(s_t,a_t)   \right]
\end{equation} \label{eq:policy}%
where $\eta$ represents the learning rate. 

Conventional Q-learning is based on tabular methods, where the observation space needs to be discretized first. There are two main practical issues: 1) The observation space discretization strategy only works well if the range and dimension of the observation space are relatively small. For large-scale problems, it easily leads to memory explosion and also requires more training time to converge a good solution; 2) The observed states in real-world environments are usually noisy or incomplete, which makes it very difficult for tabular methods to capture the true pattern based on noisy data. We tested the performance of conventional Q-learning with noisy input data in Section \ref{sec:Results}.
\vspace{-4mm}

\subsection{Deep Reinforcement Learning}

Deep reinforcement learning is a combination of RL and deep learning technologies. The use of deep learning makes it possible with DRL to directly use the raw state representations, and train policies for complex systems and tasks with effective and efficient approaches for high-dimensional feature extraction and non-linear generalization. DRL algorithms learn directly from agents' interactions with an environment (either simulation or real). Although catastrophic events rarely happen in the real world, a wide variety of extreme event scenarios can be created in simulation and provide the DRL algorithms extensive experience to learn. This is unlike other deep learning techniques that require a large amount of labelled training data, which is usually sparse or not available in the power industry. 

One of the most successful DRL algorithms suitable for discrete action space is DQN, which uses neural network (NN) with weights $\theta$ to estimate Q-values. \hl{Compared to conventional Q-learning with the function approximation approach{\cite{Sutton1998}}, which usually requires a significant amount of manual tuning to stabilize the learning process, there are two key traits that make DQN more efficient and stable}: 1) the use of a target network $(\hat{Q})$ besides the Q-network; and 2) the use of experience replay \cite{Mnih2015}.
The target network has the same structure as the Q-network. At regular periodicity (every $\tau$ steps), the weights of the Q-network are copied to the target network \cite{Mnih2015}. To perform experience replay, the agent's experience 
$e_t = (s_t, a_t, r_t, s_{t+1})$ is stored in data set $D$ at each time step. A Q-network can be trained using samples (minibatches) randomly drawn from $D$ by minimizing a sequence of loss function (\ref{eq:loss})  \cite{Mnih2015} 
\begin{equation} \label{eq:loss}
L_i(\theta_i) = E_{(s,a)\sim p}  \left[ (y_i - Q(s,a;\theta_i))^2   \right] 
\end{equation}
where $y_i$ is the target Q-value for iteration $i$ computed by $\hat{Q}$, and $p$ is the probability distribution of the state and action pair $(s,a)$. Updating NN weights θ can be done by stochastic gradient descent with the gradient calculated by (\ref{eq:5}).

\vspace{-2mm} 
\begin{equation}\label{eq:5}
\nabla_{\theta_i} L_i (\theta_i) = E_{(s,a)\sim p}  \left[ (y_i - Q(s,a;\theta_i))\nabla_{\theta_i} Q(s,a;\theta_i)  \right] 
\end{equation} 
\vspace{-3mm} 

A popular algorithm for training DQN is presented as \textbf{Algorithm 1} below \cite{Mnih2015}.
\begin{table}[!ht]
\label{tab:algo}
\raggedright
\begin{tabular}{p{0.1cm} p{7.6cm}}
\hline
 \multicolumn{2}{c}{\textbf{Algorithm 1} Deep Q-learning}   \\
\hline
1 & Initialize  $Q(s,a;\theta)$ and target network $\hat{Q}$ with random weights $\theta_0$\\
2 & Initialize experience replay memory $D$, \hl{exploration rate $\epsilon = 1$} \\
3 & \textbf{For} episode $n =1$, \textit{M} \textbf{do} \\
4 &	$\quad s\leftarrow s_1$ 	Initialize the environment	\\
5 &	$\quad$\textbf{For} $t=1$, \textit{T} \textbf{do}	\\
6 &	$\quad \quad$With probability $\epsilon$ select a random action $a_t$	\\
7 &	$\quad \quad$Otherwise select $a_t = max_a Q(s_t,a;\theta)$	\\
8 &	$\quad \quad$Execute $a_t$ in the environment and 	\\
9 &	$\quad \quad$	observe reward $r_t$ and next state $s_{t+1}$	\\
10 & $\quad \quad$Store transition $(s_t,a_t,r_t,s_{t+1})$ in $D$	\\
11 & $\quad \quad$Sample random batches $(s_j,a_j,r_j,s_{j+1})$ from $D$	\\
12 & $\quad \quad$\textbf{If} $s_{j+1}$ is a terminal state \textbf{do} \\
13 & $\quad \quad \quad$ $y_j = r_j$		\\
14 & $\quad \quad$\textbf{else}		\\
15 & $\quad \quad \quad$ $y_j = r_j + \gamma * max_a\hat{Q}(s_{j+1},a;\theta)$		\\
16 & $\quad \quad L(\theta)=\left(y_j -Q(s_t,a;\theta) \right)^2 $		\\
17 & $\quad \quad \theta\leftarrow \theta -\eta \nabla_{\theta}L(\theta)$		\\
18 & $\quad \quad$Every $\tau$ step, reset $\hat{Q} = Q$ 		\\	
19 & $\quad$\textbf{End For}			\\
20 & $\quad$\hl{If $\epsilon > \epsilon_{min}$, $\epsilon \leftarrow \varphi(n, \epsilon)$} \\
21 & \textbf{End For}		\\
\hline
\end{tabular}
\end{table}
With the implementation shown in \textbf{Algorithm 1}, DQN uses every possible data tuple and break correlation  in the observation sequence by sampling from experience replay, which benefits data efficiency and reduces training variance. \hl{The exploration rate $\epsilon$ in the state-of-the-art implementation is usually not constant, but decays (linearly in our experiments) from 1.0 to a small constant value $\epsilon_{min}$ within certain steps, which is defined as $\varphi(n, \epsilon)$ in Algorithm 1. It means that the agent will explore more in the beginning and exploit more at the end}. DQN approximates Q values based on neural networks, so it avoids the memory explosion problem caused by observation space discretization in traditional Q-learning. At last, DQN can capture the underlying pattern(s) even from noisy observations, which will be shown in   Section \ref{sec:Results}.

\hl{Note that we represent Algorithm 1 from a general perspective, with good generalization capabilities that could be adapted  for and interact with many different environments. The key steps in Algorithm 1 for interaction with a specific grid control environment are highlighted as follows:
(1) step 4---initialize the environment; (2) step 8---execute an action in the environment; (3) step 9---observe reward $r_t$ and next state $s_{t+1}$; and (4) step 12---check whether $s_{j+1}$ is a terminal state. More details of this Deep Q-learning algorithm interacting with the grid control environment will be discussed in the following sections.}

\subsection{Grid Emergency Control}
\hl{
%% add three main operation stages based on Dynclon?
% The main procedure of power grid emergency control is that during and after some large disturbance(s) in the power grid, based on the observations, the agent(s) control the actuation devices(e.g., a breaker) to make sure the power grid evolves within safety and stability margins and eventually recovers to a secure operation condition within a short time horizon ( from tens of seconds to minutes). %
For large-scale power systems, the emergency control problem is a highly non-linear, non-convex optimal decision-making problem and can be formulated as follows:
}
\begin{equation}\label{eq: emergency_control_optimal}
%\begin{align}{6}
\bf{P1:}\quad \min \int\limits_{{T_0}}^{{T_c}} {C\left( {{{\bf{x}}_t},{{\bf{y}}_t},{{\bf{a}}_t}} \right)dt}  \\
\end{equation}
\vspace{-3mm} 
\textit{s.t.}
\vspace{-3mm} 
\begin{IEEEeqnarray}{r'c'l}
{{\bf{\dot x}}_t} = f({{\bf{x}}_t},{{\bf{y}}_t},{d_t},{{\bf{a}}_t})  \IEEEyessubnumber\label{eq:subeq2}\\
0 = g({{\bf{x}}_t},{{\bf{y}}_t},{d_t},{{\bf{a}}_t}) \IEEEyessubnumber\label{eq:subeq3} \\
{\bf{x}}_t^{\min } \le {{\bf{x}}_t} \le {\bf{x}}_t^{\max }{,\quad}\forall t \in \left[ {{T_0},{T_c}} \right]  \IEEEyessubnumber\label{eq:subeq4} \\
{\bf{y}}_t^{\min } \le {{\bf{y}}_t} \le {\bf{y}}_t^{\max }{,\quad}\forall t \in \left[ {{T_0},{T_c}} \right]  \IEEEyessubnumber\label{eq:subeq5} \\
{\bf{a}}_t^{\min } \le {{\bf{a}}_t} \le {\bf{a}}_t^{\max }{,\quad}\forall t \in \left[ {{T_0},{T_c}} \right] \IEEEyessubnumber\label{eq:subeq6}  
\end{IEEEeqnarray}
\hl{where $\bf{x}_t$ represents dynamic state variables of the power grid, such as the generator rotor angles and speeds, etc.; $\bf{y}_t$ represents the algebraic state variables of the power grid, which are typically the voltages at nodes (or buses) of the grid; $\bf{a}_t$ represents the emergency control variables of the power grid, such as generator tripping or load shedding; and $d_t$ represents the disturbance (or contingency) that could occur in the grid. $T_0$ and $T_c$ represent the time horizon. $C(\cdot)$ represents the cost function of the power grid emergency control. The dynamic behavior of various components in the power grid, such as generators and their controllers, is represented by ({\ref{eq:subeq2}}). Eqn. ({\ref{eq:subeq3}}) represents the algebraic constraints that describe  the network coupling between generators, loads, and transmission branches in the power grid. Eqns. ({\ref{eq:subeq4}}), ({\ref{eq:subeq5}}) and ({\ref{eq:subeq6}}) represent the operation and security constraints on the dynamic state variables, algebraic state variables, and control variables over the time horizon. Notice that the upper and lower bounds in ({\ref{eq:subeq4}}), ({\ref{eq:subeq5}}) and ({\ref{eq:subeq6}}) could be time-variant. The emergency control problem formulated as $\bf{P1}$ can be solved by a model predictive control (MPC) method {\cite{Jin2010MPC}}}. % or cite guanji's paper.

\hl{The same problem can also be formulated as an MDP and solved by RL methods. Note that not all the state variables are observed by the agent(s); thus, the state space $\mathcal{S}$ in MDP is a subset of the grid state variables, i.e.,  $s_t \subset \{\bf{x_t} \cup \bf{y_t}\}$. It should be noted that properly defining $\mathcal{S}$ for specific emergency control problems is critical. Two specific examples, along with general design principles, will be discussed in section {\ref{sec:Algorithm}}. Based on the properties of the control actions $\bf{a_t}$, an action space $\mathcal{A}$, either continuous or discrete, will be defined. The limits on the controls defined in ({\ref{eq:subeq6}}) are generally considered in the definition of the action space by setting the bounds. The environment transition from $s_t$ to ${s_{t+1}}$ (i.e., steps 8 and 9 in Algorithm 1) is governed by the differential and algebraic equation set ({\ref{eq:subeq2}}) and ({\ref{eq:subeq3}}). The detailed formulations of ({\ref{eq:subeq2}}) and ({\ref{eq:subeq3}}) and the solution methods can be found in {\cite{kundur1994power}}. The reward $r_t$ is a function of $\bf{x_t}$, $\bf{y_t}$ and $\bf{a_t}$ as follows:}
\begin{equation} \label{eq:reward}
 	{r_t} = h({{\bf{x}}_t},{{\bf{y}}_t},{{\bf{a}}_t}) 
\end{equation}
\hl{where $h(\cdot)$, in principle, should incorporate both the action cost function $C(\cdot)$ in ({\ref{eq:subeq3}}) and a penalty of any violation of the constraints defined in ({\ref{eq:subeq4}}), ({\ref{eq:subeq5}}) and ({\ref{eq:subeq6}}). Detailed formulations of $h(\cdot)$ for two specific emergency control schemes will be presented in Section {\ref{sec:Algorithm}}.}

%%%%%%%%%%%%%%%%%%%%%%%%%%%%%%%%%%%%%%%%%%%%%%%%%%%%%%%%%%%%%%%%%%%%%%%%%%%%%%%%%%%%%%%%%%%%

\section{An Open Platform for Developing and Benchmarking RL Algorithms for Grid Control}\label{sec:Platform}

\subsection{Overview}
 An open-source platform, Reinforcement Learning for
Grid Control (RLGC), has been developed and published
for the purpose of developing, training and benchmarking
RL algorithms for power system control \cite{RLGC}. \hl{Open-source benchmarks (such as ImageNet and OpenAI Gym) are the key driving forces that propel the advancement of machine learning (including RL). The goal of RLGC is to create a similar open-source benchmark for reinforcement learning for power grid control.} 

\begin{figure}
\vspace{-4mm} 
\centering
\includegraphics[width=0.45\textwidth]{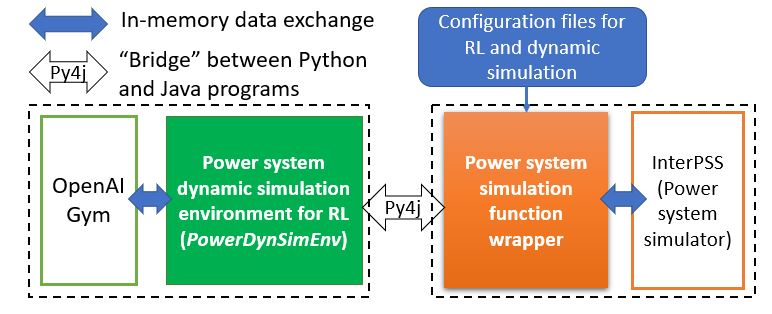}
\caption{An open platform for developing, training and benchmarking RL algorithms for power system control }
\label{fig:1}
\vspace{-5mm} 
\end{figure}

The architecture of this open platform is shown in Fig. \ref{fig:1}. It has two main modules: 1) the RL module; and 2) the power system simulation and control module. The RL module is developed based on OpenAI Gym, which is a widely-used generic toolkit for RL research and is programmed in Python  \cite{OpenAI}. A general power system simulation and control environment for training and testing RL algorithms is created, where the power system simulation and control module  is called. The power system simulation and control module is developed based on InterPSS \cite{Zhou2017} and programmed in Java. Both modules are decoupled and communicated through Py4J \cite{Py4J}, which acts as a communication ``bridge'' between Python and Java programs. The data exchange through Py4J  between the two modules is in-memory, with high efficiency and integration flexibility. Two configuration files are used to specify the power system dynamic simulation settings and the RL training parameters, respectively. 
%\vspace{-4mm} 

One main advantage of choosing OpenAI Gym for the platform is that users can directly use state-of-the-art open-source learning algorithms such as OpenAI Baselines \cite{Baseline}, which is a set of high-quality implementations of DRL algorithms, \hl{such as DQN for  MDPs with discrete control actions  and PPO  {\cite{Schulman2017}} for MDPs with continuous control actions.} We use the DQN implementation in OpenAI Baselines for solving two  emergency control control problems with discrete control spaces in this paper. 

With a modular, decoupled architecture design, as well as open-source tools   adopted for its development, the RLGC platform is:

1) \emph{Extensive}: the framework can capture many diverse aspects of RL and power systems, such as abundant choices of different RL training algorithms, rich power system dynamics and measurements, and typical emergency control actions. It can also simulate various power systems, including integrated transmission and distribution systems \cite{Huang2017}. 

2) \emph{Flexible}: With this platform, users only need to specify a minimum of two configuration files   to build a customized environment for training and testing RL algorithms for power system control. Users can define various observations, actions and rewards through either a configuration file or programming new functions for them.

\subsection{Implementation Details and Usage}

In the RL module, a python class named \textit{PowerDynSimEnv} is developed by extending the OpenAI Gym's standard basic environment \textit{Env} class. 
In the power system simulation and control module, a wrapper of InterPSS simulation functions and capabilities is developed for interfacing with the \textit{PowerDynSimEnv}  environment in the RL module.  \hl{It comprises several key functions representing the interactions between the learning agent and the environment in   Algorithm 1 (AL1). The key functions include  \textit{initStudyCase(*)} for   initializing the environment of AL1 in step 4, \textit{applyAction(*)} and \textit{nextStepDynSim(*)} for executing action $a_t$ in step 8, \textit{getReward(*)} and \textit{getEnvObversations(*) } for observing reward $r_t$ and the next state $s_{t+1}$, and \textit{isSimulationDone(*)} for checking if $s_{j+1}$ is a terminal state of AL1.} The usage of these functions for RL training will be detailed in the following paragraph.

A typical procedure for using the developed platform to test DRL algorithms and train NN models for grid control  \hl{mainly includes two stages: (1) the training stage for learning, and (2) the testing stage for validating the trained NN. During the training stage, the DRL will perform neural network learning through a large number of training steps. It learns    an optimal policy with exploration and exploitation, and  automatically saves the best-performance NN parameters. Once the training stage is completed, the RL agent at the testing stage will   use the learned optimal policy   (represented by the best-performance NN parameters)  to provide optimal control actions to the environment, based on the observed environment states. }

\begin{figure}
\centering
\includegraphics[width=0.42\textwidth]{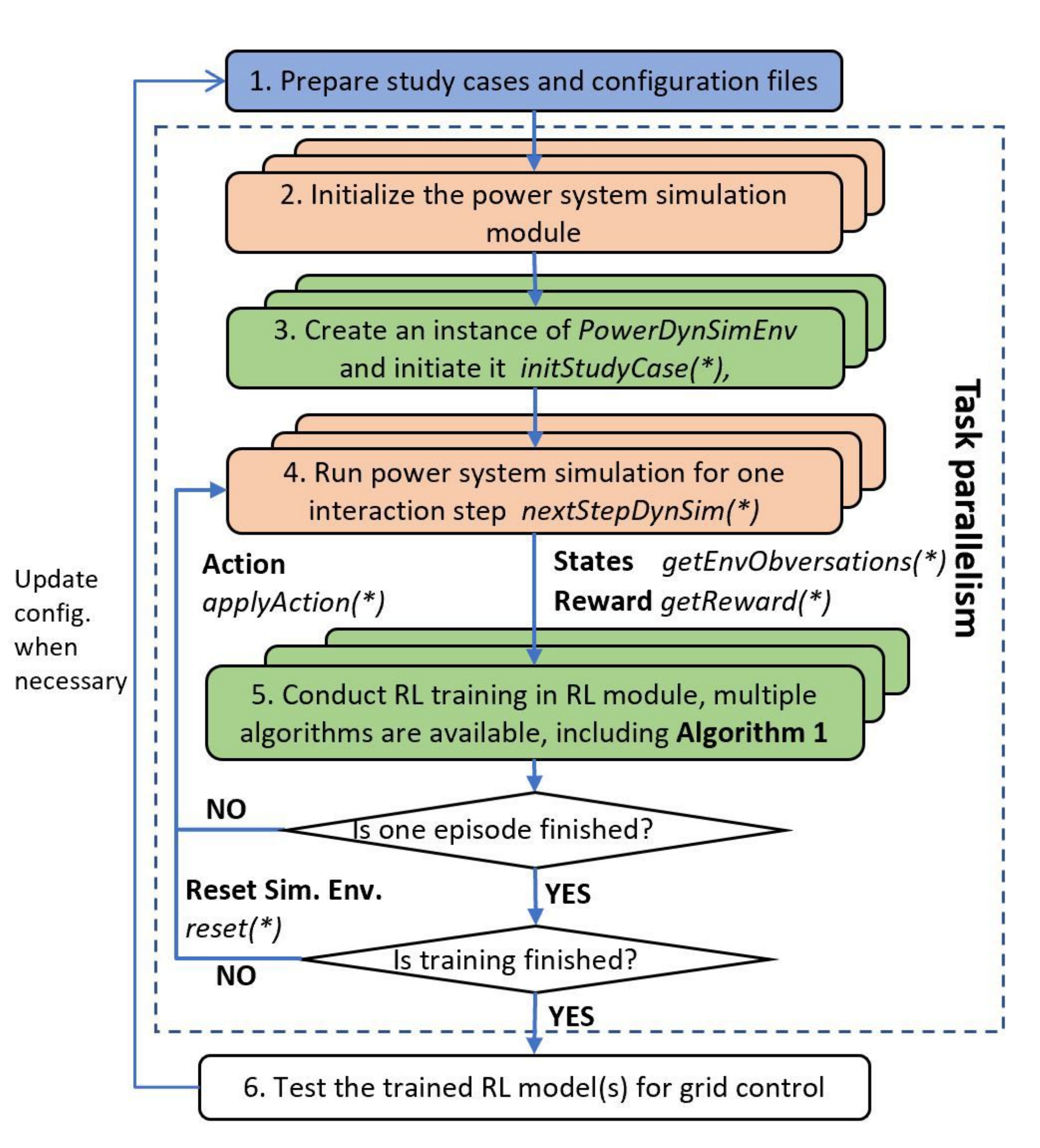}
\caption{A flowchart illustrating a typical procedure for using the platform for training and testing RL model(s) for grid control }
\label{fig:2}
\vspace{-3mm} 
\end{figure}

\hl{Fig. {\ref{fig:2}} gives the details of the procedure for using the platform for training and testing the DRL model for grid control. }
Once the study cases and configuration files described in Section \ref{sec:Platform}.A are prepared, the training procedure initializes power
system simulation module \textit{(initStudyCase(*))}, the NN
model, the RL module, and then launches the training. At each training step, the agent in the RL module receives the states \textit{(getEnvObversations(*))} and rewards \textit{(getReward(*))} from the environment, which calls the power system simulation module to obtain these information, trains the NN model (see Algorithm 1 for details of training algorithm), and sends back the selected control action to the simulation environment. Upon receiving the control action from the RL module, the power system simulation module applies this control action in the dynamic simulation\textit{(applyAction(*))}, runs to the next agent-environment interaction step \textit{(nextStepDynSim(*))}, and sends the updated states and rewards to the RL module. It should  also be noted that power system dynamic simulation has its own time step (ranging from 1 ms to half cycle) to ensure numerical stability and which is usually smaller than the time step of the DRL module (agent) interacting with the power system simulation module (environment); thus, there is an internal power system simulation loop within \textit{nextStepDynSim(*)} function. These interactions between the two modules continue until the training reaches the end of one dynamic simulation  as one training episode finishes. At the end of each training episode, the training procedure re-initializes the dynamic simulation \textit{(reset(*))} and starts the next training episode. The training procedure ends after a predefined number of training steps. Once the training is finished, the trained NN model could be tested for cases different from the training cases to validate the effectiveness of the training. Based on the testing results, the users may adjust the training parameters and case settings in the configuration files and launch more training tasks. To facilitate the training process, the platform supports task-level parallelism, so that multiple RL training tasks with different hyper-parameters can be run in parallel.

%%%%%%%%%%%%%%%%%%%%%%%%%%%%%%%%%%%%%%%%%%%%%%%%%%%%%%%%%%%%%%%%%%%%%%%%%%%%%%%%%%%%%%%%%%%%

\section{DRL Algorithms for Grid Emergency Control}\label{sec:Algorithm}

With the developed platform discussed in the previous section, we investigated and developed DRL-based control schemes for two typical types of grid emergency control: 1) generator dynamic brake \cite{Ernst2004}; and 2) under-voltage load shedding. In the following subsections, the DRL algorithm design  and implementation details for both emergency control schemes, including neural networks, observations, actions, and rewards will be discussed.
\vspace{-3mm}

\subsection{Neural Network Architecture}
\begin{figure}
\centering
\includegraphics[width=0.40\textwidth]{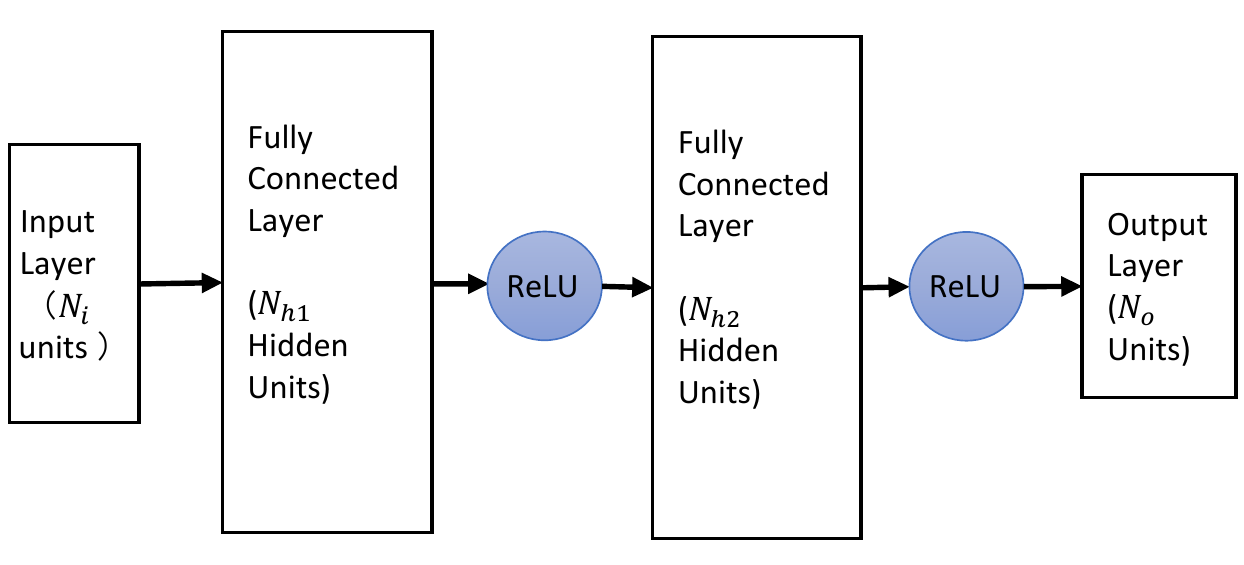}
\caption{The architecture of the NN for the grid emergency control RL agent}
\label{fig:3}
\vspace{-4mm} 
\end{figure}

The proposed architecture of the NN for both emergency control schemes is shown in Fig. \ref{fig:3}.  The number of units in the input and output layers are $N_{i}$ and $N_{o}$. There are two hidden layers in between, with $N_{h1}$ and $N_{h2}$ hidden units, respectively, which are followed by a rectified linear unit (ReLU). \hl{It should be noted that there seems to be a misconception that NNs in DRL methods have to be ``deep'' to make them work well. In fact, the groundbreaking DRL application in {\cite{Mnih2015}} and most of the DRL algorithms in OpenAI Gym continuous control benchmarks {\cite{Brockman2016}} use NNs with 2-3 hidden layers. In the reinforcement learning domain, the term ``deep'' often means a set of recent approaches that makes it possible to train a NN model using reinforcement learning, such as target network, replay buffer, duel network, etc.} The fact that the same or very similar NN architecture can be used for significantly different control problems is one main advantage of DRL over traditional RL methods like Q-learning.

\vspace{-3mm}

\subsection{Generator Dynamic Brake}
% consider to highlight observation space, action space, rewards... with 1) \emph{Reward function}:
Generator dynamic brakes are utilized to achieve two main objectives :1) to avoid the loss of synchronism between the generators when a severe incident occurs; and 2) to damp large electromechanical oscillations \cite{Ernst2004}. Due to the energy losses and operation limits,  the time the dynamic brake is switched on is limited; thus, it should be used only under emergency conditions. To achieve these objectives under the operational constraints, the following reward function \cite{Ernst2004} is used:
\begin{equation}
 r(x,u)= 
\begin{cases}
     -|\omega| -cu ,& \text{if } |\delta| \leq \pi \text{ rad } \\
    -1000,              & \text{otherwise}
\end{cases}
\end{equation}
where $\omega$ and $\delta$ are the average generator speed and angle defined in \cite{Pavella2012}, $u$ denotes the control action ($u=0$ when the brake is switched off and $u=1$ when it is switched on)  and $c$ is a penalty factor for penalizing the brake action. When the system has lost synchronism (when $|\delta| > \pi$  rad in this paper), a very negative reward (-1000) is given to direct the agent to perform appropriate actions to avoid such conditions. 

Unlike using pseudo-states (i.e., generator equivalent angle and speed) as observations in the previous research effort with RL \cite{Ernst2004}, which is essentially a hand-crafted feature extraction process, the rotor angles and speeds of monitored generators are directly used as the observation for the agent (input to the NN) in the proposed scheme. Note that it is impossible for the agent to learn the system's dynamic behaviors and the trend solely based on current observed states $\mathbf{O}_t$. Similar to stacking $m$ most recent frames as input in \cite{Mnih2015}, a sequence of observations (the number is $N_r$) is treated as a distinct state in this paper, i.e., $s_t =(\mathbf{O}_{t-N_r-1},\cdots, \mathbf{O}_t )$. In the developed platform, the number of measurements (i.e., $N_m$) and $N_r$ are configurable and defined by the users, thus $N_i  = N_m \times N_r$.

\vspace{-3mm}
\subsection{Under-voltage Load Shedding}
\begin{figure}
\centering
\includegraphics[width=0.42\textwidth]{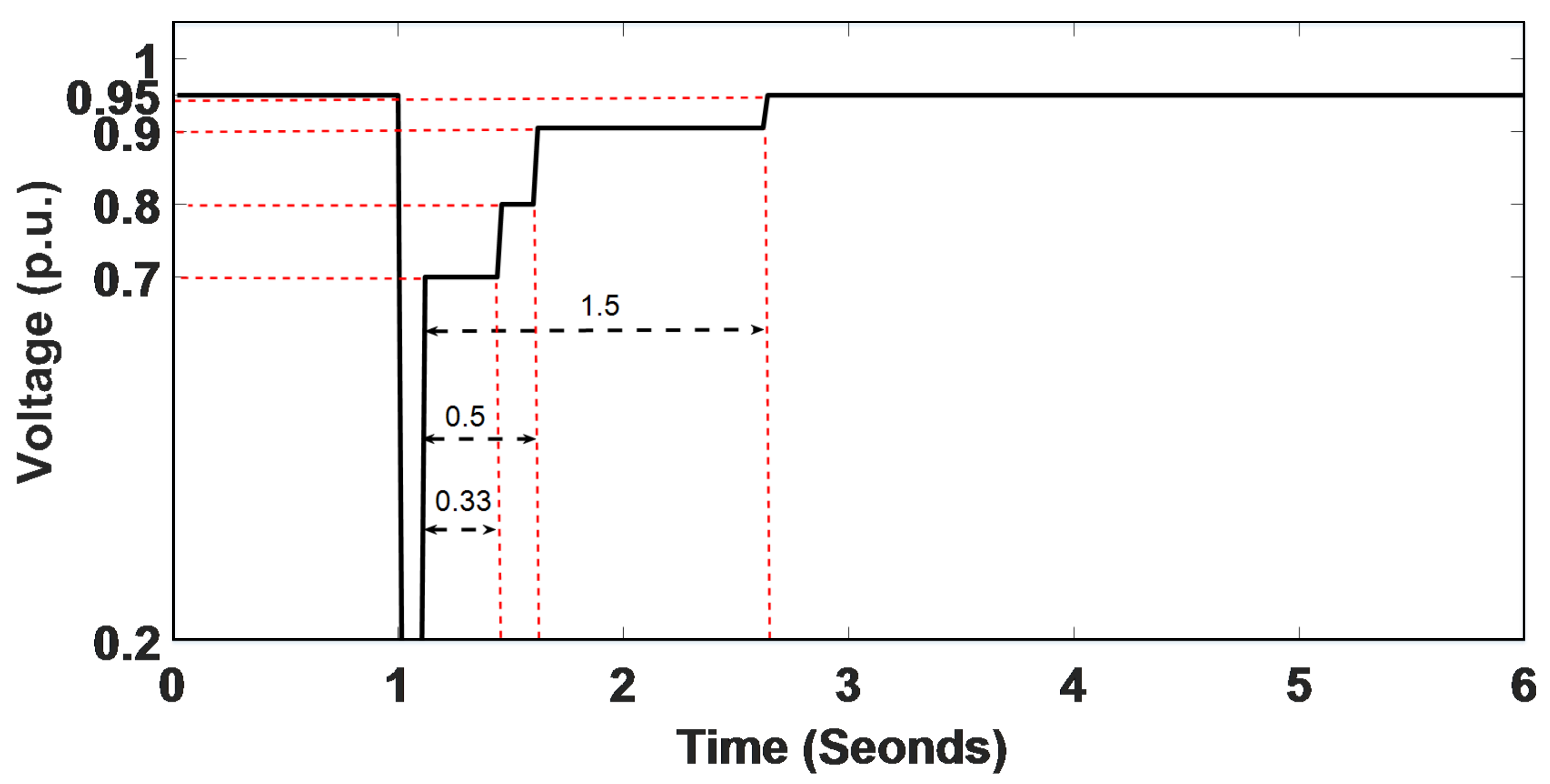}
\caption{Transient voltage recovery criterion for transmission system \cite{PJM2009}}
\label{fig:4}
\vspace{-6mm} 
\end{figure}

Fault-induced delayed voltage recovery (FIDVR) is defined as the phenomenon whereby system voltage remains at significantly reduced levels for several seconds after a fault has been cleared \cite{NERC2009}. The root cause is stalling of residential air-conditioner (A/C) motors and prolonged tripping. FIDVR events occurred in many utilities in the US. Concerns over FIDVR issues have increased since residential A/C penetration is at an all-time high and growing rapidly. A transient voltage recovery criterion (TVRC) is defined to evaluate the system voltage recovery. Without loss of generality, we referred to the standard proposed in \cite{PJM2009} and shown in Fig. \ref{fig:4}. After fault clearance, the standard requires that voltages should return to at least 0.8, 0.9 and 0.95 p.u. within 0.33 s, 0.5 s and 1.5 s, respectively. \hl{Per current industry practice, UVLS relays are usually employed to shed load demands at substations in a step-wise manner if the monitored bus voltages fall below the predefined voltage thresholds to protect power systems against FIDVR.
The ULVS relay has a fast response, however, this
distributed control scheme does not have any communication
or coordination between other substations, thus, it could lead
to unnecessary load shedding  {\cite{Bai2011}} at affected substations.
MPC methods {\cite{Jin2010MPC}}{\cite{MPC2}}  have   been proposed for UVLS protection. The MPC methods utilize a system model (usually in the form of differential algebraic equations) to predict the states of the power grid. It formulates and solves an optimization problem to decide load shedding control actions. MPC is a centralized method and   considers the coordination of load shedding between different substations. However, the optimization process in MPC methods is usually computationally intensive, and the performance of MPC methods heavily depends on the accuracy of the system model {\cite{Jin2010MPC}}. }
In this paper, we investigated applying DRL to multiple load-serving substations to implement an adaptive, coordinated emergency load shedding scheme against FIVDR. 

The observed states $\mathbf{O}_t$ at time $t$ include voltage magnitudes at monitored buses (denoted as $V_t$), as well as the percentage of load still remaining at controlled buses (denoted as $P_{Dt}$). To capture the dynamics of the voltage change, the most recent observed states are stacked with some history state records and treated as the input into DQN at time $t$, i.e., $s_t =(\mathbf{O}_{t-N_r-1},\cdots, \mathbf{O}_t )$.
The control action at each controlled load bus is defined as either 0 (no load shedding) or 1 (shed 20\% of the initial total load) at each action time step. Thus the control action space is discrete with a dimension of $2^n$, where $n$ is the number of controlled buses. The reward $r_t$ at time $t$ is defined as follows:
\begin{equation}\label{eq: UVLS_reward}
 r_t= 
\begin{cases}
    -1000, \text{ if } V_i(t)<0.95, \quad t>T_{pf} +4		 \\
    c_1 \sum_i \Delta V_i(t) -c_2 \sum_j \Delta P_j (p.u.) -c_3u_{ivld},             \text{ otherwise}
\end{cases}
\end{equation}
\[
\Delta V_i(t)= 
\begin{cases}
    min \left\lbrace V_i(t) - 0.7, 0 \right\rbrace , \text{ if }  {\scriptstyle T_{pf}<t<T_{pf} +0.33	}	 \\
    min \left\lbrace V_i(t) - 0.8, 0 \right\rbrace , \text{ if }  {\scriptstyle T_{pf} + 0.33<t<T_{pf} +0.5	}\\
    min \left\lbrace V_i(t) - 0.9, 0 \right\rbrace , \text{ if }  {\scriptstyle T_{pf} + 0.5<t<T_{pf} +1.5}	\\
    min \left\lbrace V_i(t) - 0.95, 0 \right\rbrace , \text{ if } {\scriptstyle  T_{pf} + 1.5<t}
\end{cases}
\]
where $T_{pf}$ is the time instant of fault clearance. The above reward function has three parts: (1) total bus voltage deviation below the standard voltage thresholds shown in Fig. \ref{fig:4}, where $V_i(t)$ is the bus voltage magnitude for bus $i$ in the power grid; (2) total load shedding amount, where $\Delta P_j (t)$ is the load shedding amount in p.u. at time step $t$ for load bus $j$; (3) invalid action penalty $u_{ivld}$  if the DRL agent still provides load shedding action when the load at a specific bus has already been shed to zero at the previous time step   when the system is within normal operation.   $c_1, c_2,$ and $c_3$ are weight factors for the above three parts. Note that the reward function will be set to a  large negative number (-1000) if any bus voltage is  below 0.95 p.u. 4 s after the fault is cleared.
\hl{Please  note that tuning the reward function is a challenge for DRL. It requires a combination of heuristics based on prior knowledge and some automated parameter search (trial-and-error selection). Here we provide some basic principles for reward function design: (a) use prior knowledge about the problem to identify a rough range for the parameters ($c_1, c_2$ and $c_3$) with regard to the proper reward values. A well-designed reward function should give higher reward values for better system performance. In this paper, we roughly estimate  the range of parameters by performing the power grid dynamic simulation by directly applying uniformly distributed actions from the defined action space; (b) once the rough ranges for the parameters are identified, randomly select several points from those ranges, then train the DRL model using the selected combination of parameters and choose the combination that performs best.}

%%%%%%%%%%%%%%%%%%%%%%%%%%%%%%%%%%%%%%%%%%%%%%%%%%%%%%%%%%%%%%%%%%%%%%%%%%%%%%%%%%%%%%%%%%%%

\section{Test Results}\label{sec:Results}

In this section, test cases and results are presented for the two typical grid emergency control schemes: 1) generator dynamic brake; 2) under voltage load shedding we discussed in Section \ref{sec:Algorithm}. \hl{All the case studies including training and testing were performed in a simulation environment (off-line mode) based on the RLGC platform.}

\vspace{-3mm} 

\subsection{Generator Dynamic Brake}
\begin{figure}[ht]
\centering
\includegraphics[width=0.44\textwidth]{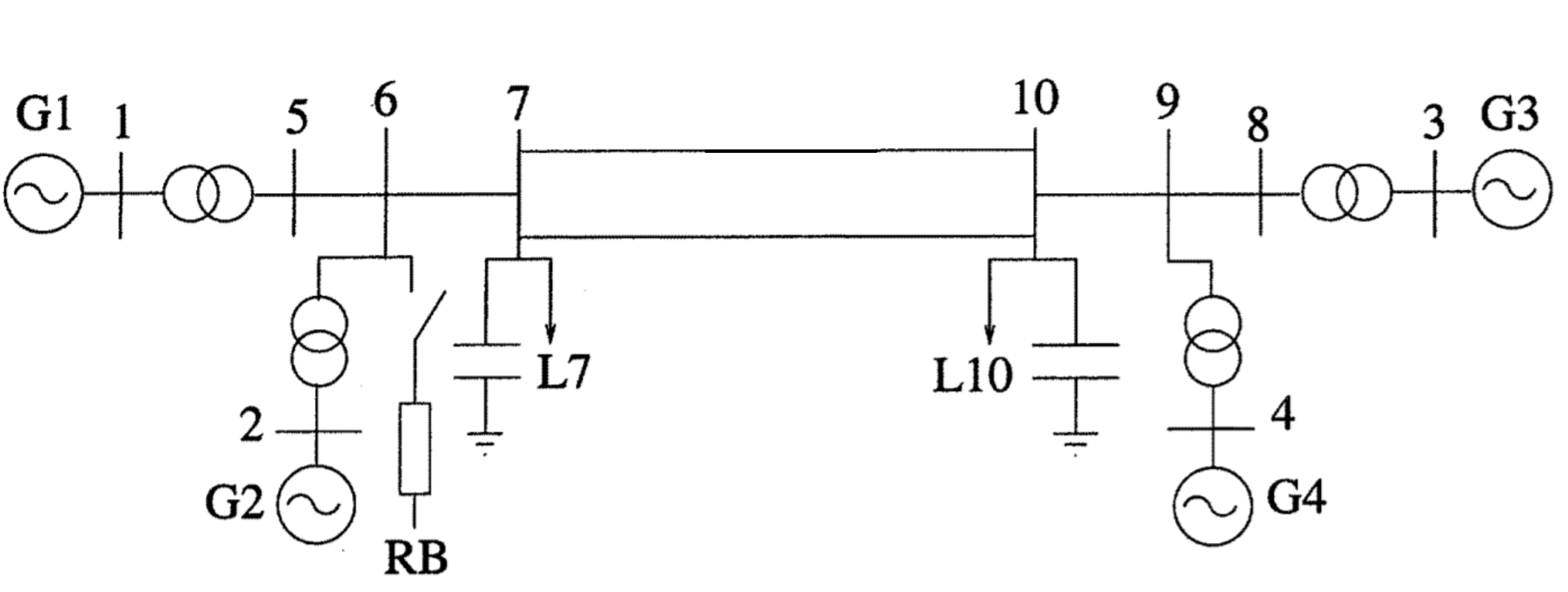}
\caption{The two-area, four-machine system with resistance brake at bus 6}
\label{fig:5}
\vspace{-3mm} 
\end{figure}

To illustrate the capabilities of the proposed DRL framework and algorithm, a generator dynamic brake  controlled by an RL agent is tested on the two-area, four-machine system, as shown in Fig. \ref{fig:5}, where the resistive brake (RB) is located at bus 6 with the size of $g=4.0$ p.u. mhos on a 100 MVA base (400 MW). The test case is very similar to the first test case in \cite{Ernst2004}. 
\vspace{-3mm} 

The observation states are the speed and rotor angles of four generators; thus, $N_m$  = 8. The last 4 recent observation states are used as input for DQN; thus, $N_r$= 4, and the number of nodes in NN input layer $N_i$ is 32. The number of nodes in the output layer $N_o$  is 2 (representing 0 and 1). Other important hyperparameters are as follows: the coefficient $c$ in (6) is 2; total interaction steps in training is 900,000;	nodes in hidden layers: $N_{h1}= N_{h2}= 128$;	learning rate $ \eta = 0.0001$; \hl{minimum exploration rate $ \epsilon_{min} = 0.02$}.

\begin{figure}[ht]
\centering
\includegraphics[width=0.32\textwidth]{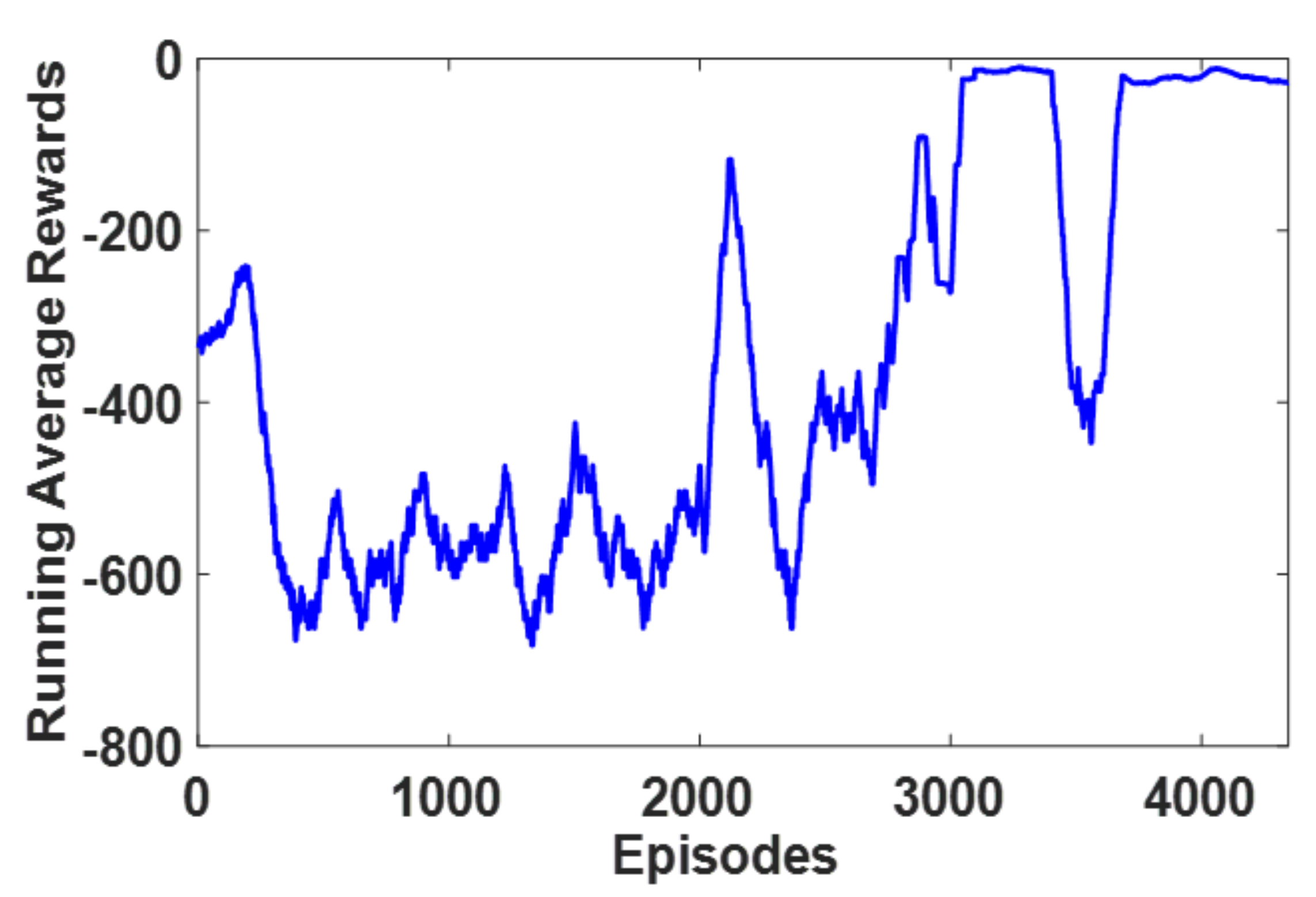}
\caption{The moving average rewards during the DRL training}
\label{fig:6}
\vspace{-2mm} 
\end{figure}

The training period is partitioned into different episodes (scenarios). Each episode begins with a flat start of dynamic simulation, and a three-phase, short-circuit fault is applied at bus 3 at 1.0 s with a random fault duration ranging from 0.581 s to 0.585 s; thus, the fault is self-cleared. This random selection of the fault duration could guarantee that the training agent interacts with both stable and unstable post-fault conditions, as the critical clearing time for three-phase faults at bus 3 is 0.583 s. For each episode, the simulation proceeds until either instability is detected or the simulation time reaches 30 s. The power system dynamic simulation time step is 0.002 s. During each episode, the agent interacts with the simulated power system environment at the time step of 0.1 s. The same time steps are used in the test cases in the rest of the paper.
\hl{It took 9 hours in a Linux workstation with 32 AMD Opteron 1.44 GHz Processors and 64 Gigabit memory with no parallelism to complete the training process.} With well-tuned parameters, our approach robustly learns successful policies. The moving average of the reward during the training is shown in Fig. \ref{fig:6}. \hl{The dip around the 3600th episode shown in Fig. {\ref{fig:6}} is corresponding to a large negative reward due to one ``bad'' exploration during training. However, this does not imply the instability of the DQN algorithm. As the training of DQN algorithm continues, the DQN model learns to avoid the bad control actions experienced in the training and converges to a local optimal solution. Extensive tests show that all the local optimums that we achieved are good solutions.}

After the DRL model training, we assess robustness of the resulting control policy (law) on a different and much larger set of scenarios, with different combinations of power flow condition, fault location, and fault duration:

1) different power flow conditions are tested, including (a) the original power flow case for training and learning, (b) each load in the system increases/decreases by 50 MW, 100 MW, and 180 MW; (c) the tie-line (two lines between buses 7 and 10 ) power flow increases/decreases by 20MW, 40 MW, 70 MW and 100 MW. \hl{Because the two tie-lines are the only connection   between area 1 and 2, the adjustment of tie-line power flow could be achieved by increasing the real power output of the generators at one area while decreasing the real power output of the generators at the other area accordingly;}

2) the fault location is selected for all the 10 buses;

3) and the fault duration is randomly selected between 0.3 s and 0.7 s.

Without the dynamic breaking, the maximum fault duration that the two-area power system can withstand without losing stability is 0.583 s. \hl{On the other hand, when the RB is used with the control law trained by DRL, for the above discussed different scenarios (we test 220 different scenarios), the system can remain stable.} To make the inputs of  the DRL-based control more realistic, we also add zero mean, 1\% Gaussian-distributed noise to the observations fed into the trained NN. We also compared the trained DRL-based control versus the conventional 2-dimension Q-table-based Q-learning method in \cite{Ernst2004}. \hl{The results show that the DRL-based control outperforms the conventional Q-learning-based control for all testing scenarios with noises added into the observations.}

\begin{figure}[!ht]
\centering
\vspace{-4mm}
\subfigure[]{\includegraphics[width=0.23\textwidth]{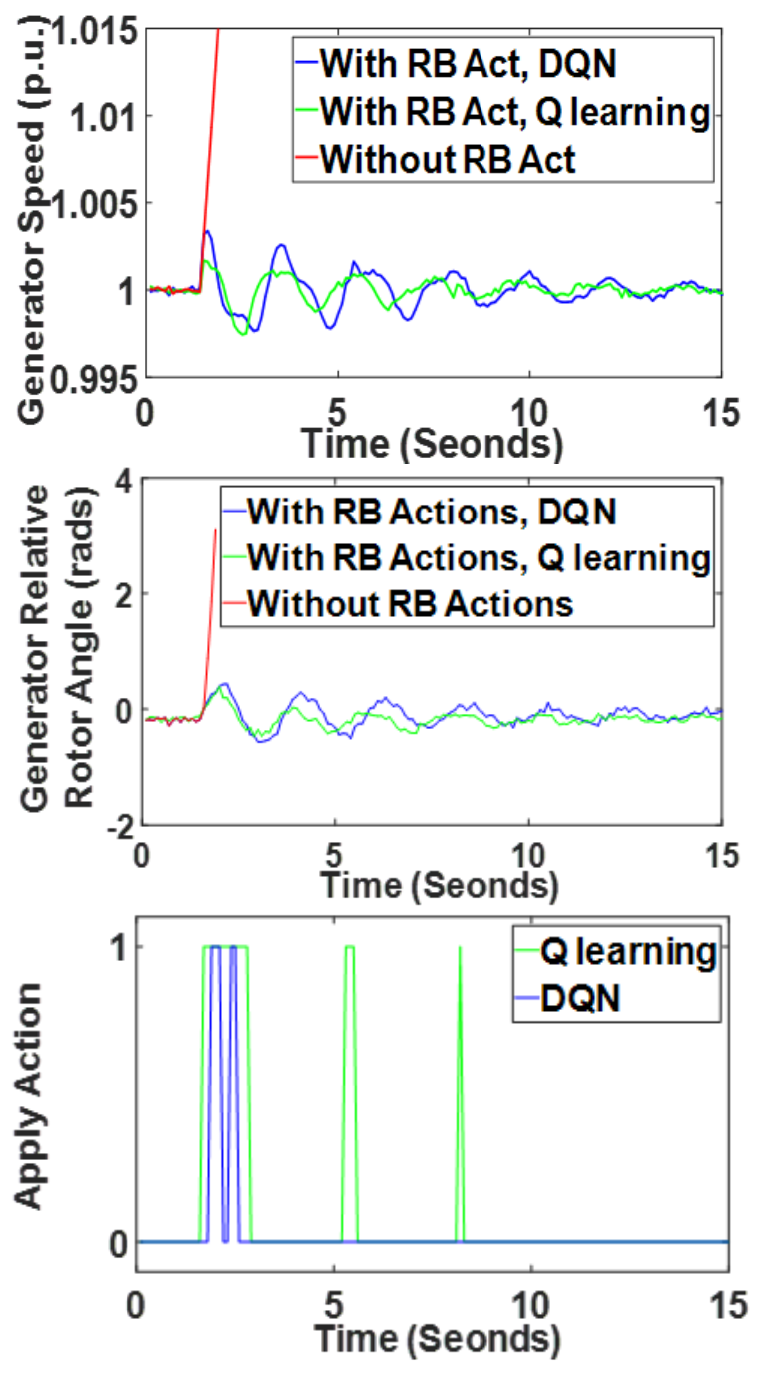}}
\subfigure[]{\includegraphics[width=0.235\textwidth]{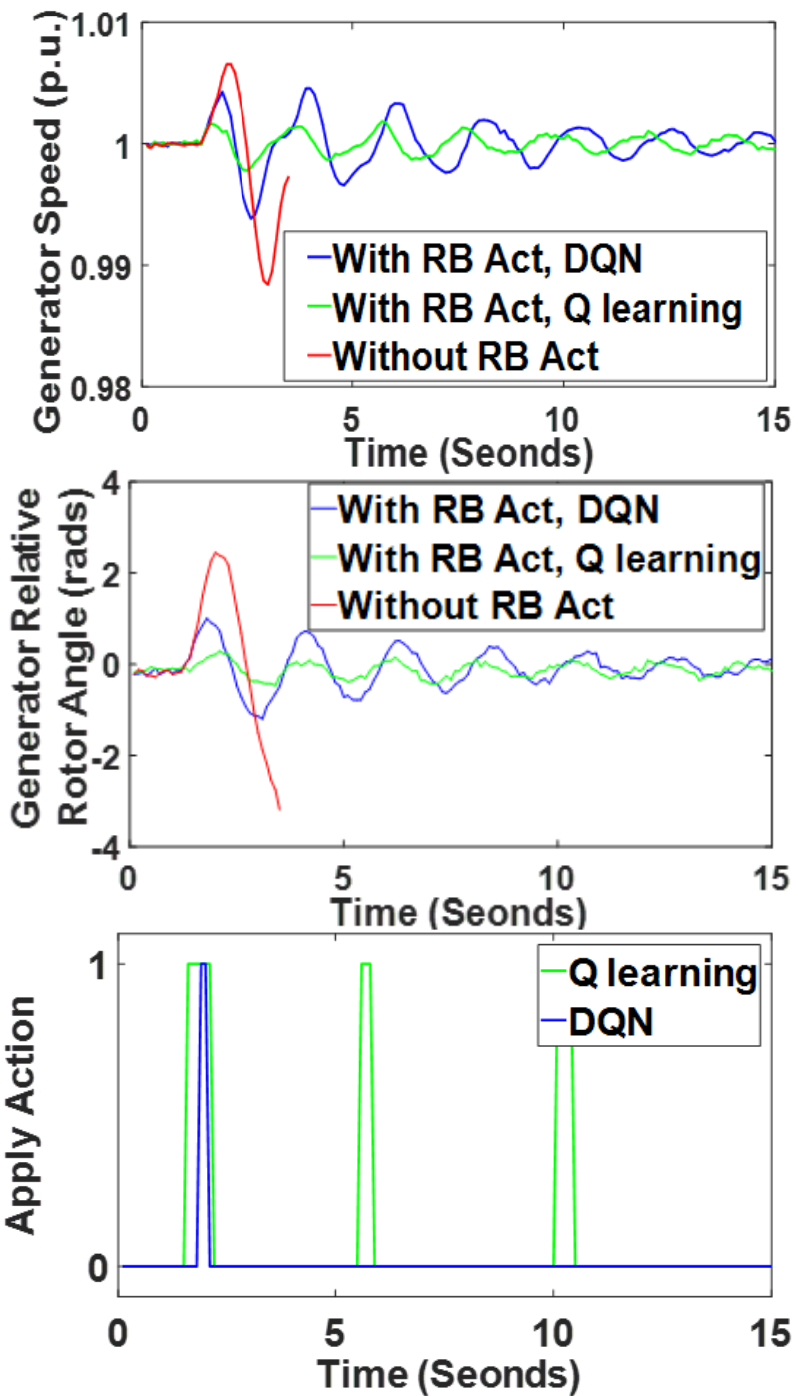}}
\caption{Evolution  of the generator speed and the system relative rotor angle: (a) 0.7 seconds fault at bus 4, heavy load power flow condition; (b) 0.6 seconds fault at bus 9, normal load power flow condition}
\label{fig:7}
\vspace{-4mm} 
\end{figure}

Fig. \ref{fig:7} (a) and (b) show two examples of the RB actions for different faults and power flow conditions, for both DRL-based and conventional Q-learning-based control. Fig. \ref{fig:7} (a) shows the generator 3 speed and the relative rotor angle (with and without RB actions), as well as the RB actions for a fault at bus 4 with a duration of 0.7 seconds, under the power flow condition that each load increases 100 MW with reference to the power flow case in the training. Fig. \ref{fig:7} (b) shows the generator 3 speed and the relative rotor angle, as well as the RB actions for a fault at bus 9 with a duration of 0.6 seconds, under the original power flow condition for training. It could be observed from Fig. \ref{fig:7} (a) and (b) that the system loses stability if there are no   RB actions (red line), while the RB actions provided by both the DRL-based (blue line) and conventional Q-learning-based control (green line) can sustain the system stability. However, the DRL-based control definitely provides better control actions than the conventional Q-learning-based control, as the DRL-based control operates the RB in less time steps and thus obtains higher rewards. It could  also be observed from Fig. \ref{fig:7} (a) and  (b) that the DRL-based control will provide different RB actions at different times for the two different scenarios. All the results shown in Fig. \ref{fig:7} demonstrate the effectiveness, robustness, and adaptiveness of the DRL algorithm. It should be noted that we also tested various pre-fault periods; the DRL-based control does not apply any braking action on the system under normal conditions.

\subsection{Under Voltage Load Shedding}
\begin{figure}[ht]
\centering
\includegraphics[width=0.37\textwidth]{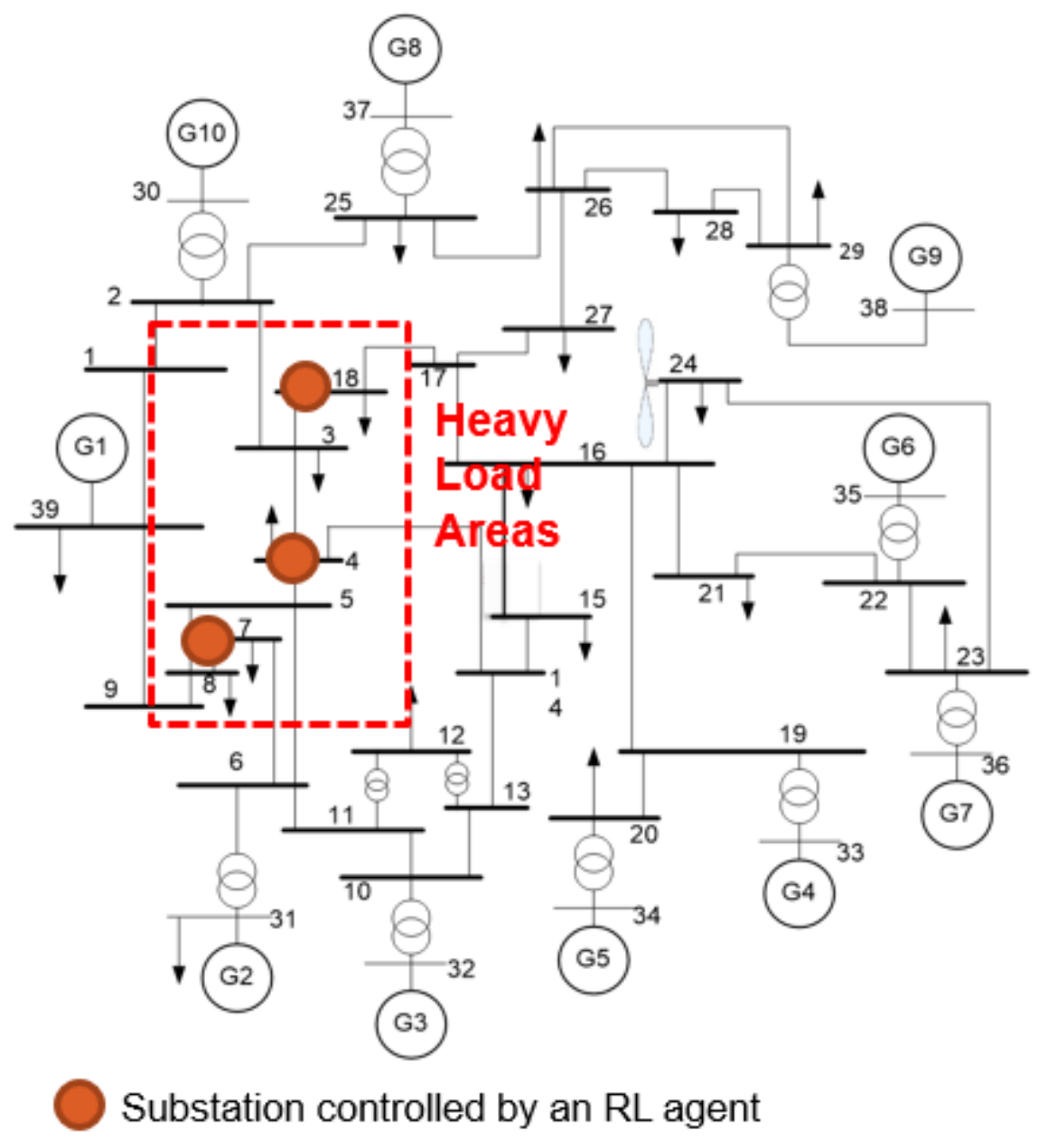}
\caption{A modified IEEE 39-bus system}
\label{fig:8}
\vspace{-3mm} 
\end{figure}

The developed platform and DRL algorithm was applied for developing a coordinated UVLS scheme against FIDVR and  was tested on a modified IEEE 39-bus system \cite{Pyo2010}, as shown in Fig. \ref{fig:8}, where step-down transformers are added to load buses 4, 7, and 18. The original loads are moved to the low-voltage side of the transformers and modelled as a combination of 50\% single-phase air-conditioner motors \cite{Kosterev2008} and 50\% constant impedance loads.

The OpenAI Baselines implementation of the DQN algorithm is used to learn a closed-loop control policy for applying the load shedding at   buses 4, 7 and 18 to avoid the FIDVR and meet the voltage recovery requirements shown in Fig. \ref{fig:4}. The coefficients of the reward function (\ref{eq: UVLS_reward}) for this study are:  $c_1= 260$, $c_2= 150$, and  $c_3= 3$.  The observations include voltage magnitudes at buses 4, 7, 8, and 18 and low-voltage sides of the step-down transformers connected to them, as well as the fractions of loads served by buses 4, 7, and 18; thus, $N_m$  = 11. The last 10 recent observation states are stacked and used as input for DQN; thus, $N_r$= 10, and the number of nodes in the NN input layer $N_i$ is 110. The control action for buses 4, 7, and 18 at each action time step is either 0 (no load shedding) or 1 (shedding 20 \% of the initial total load at the bus). Thus, the total number of combinations of potential discrete control actions at each action step is 8, i.e., the number of nodes in the output layer $N_o$ is 8. Other important hyperparameters are as follows: 	total interaction steps in training is 1,200,000; nodes in hidden layers $N_{h1}= N_{h2}= 256$; 	learning rate $\eta= 0.00005$;  minimum exploration rate $ \epsilon_{min} = 0.02$.
 
\begin{figure}[ht]
\centering
\includegraphics[width=0.36\textwidth]{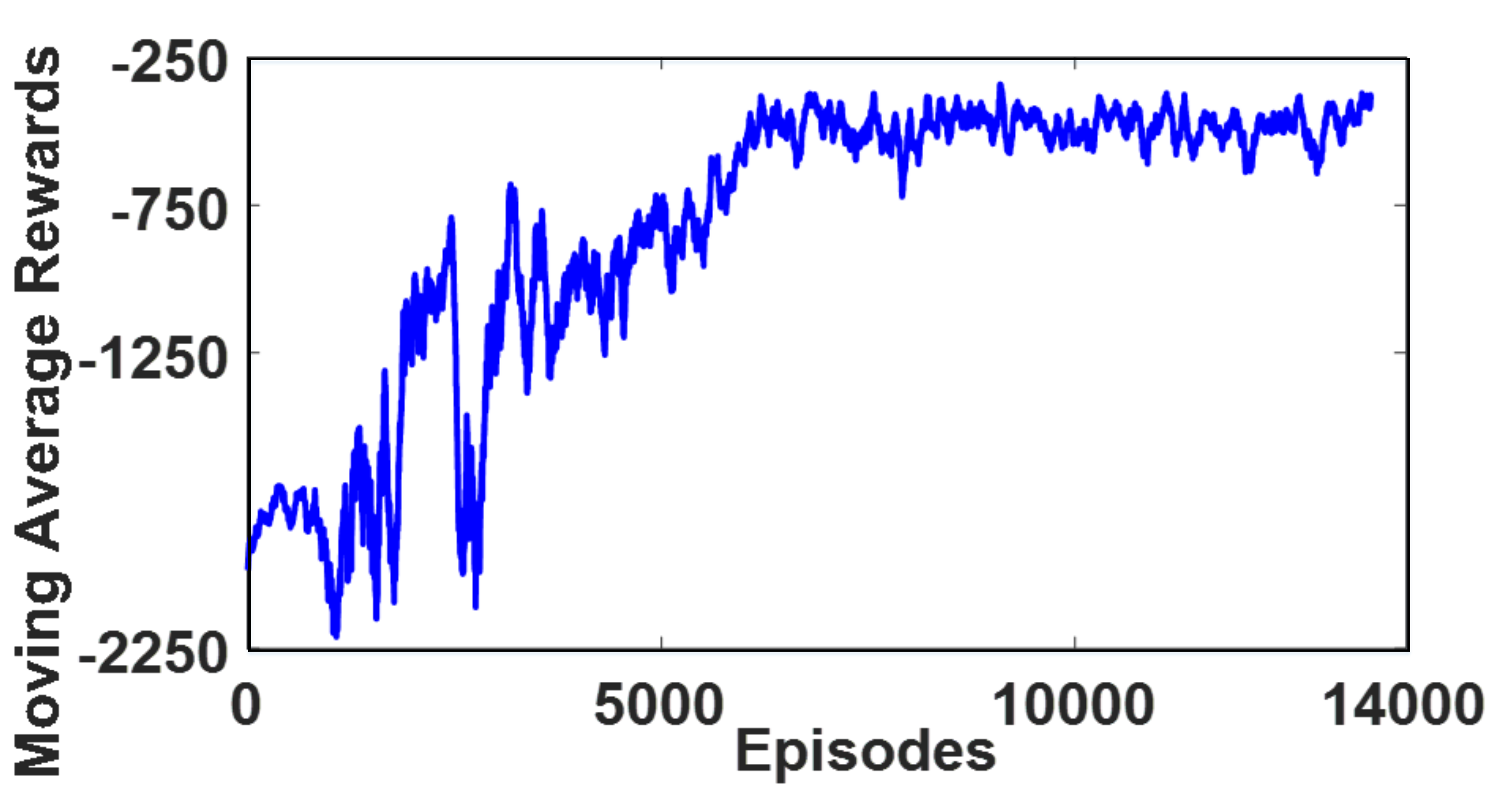}
\caption{The moving average of the rewards during the DRL training for load shedding control in the 39-bus system}
\label{fig:9}
\vspace{-2mm} 
\end{figure}

 During the training, each episode begins with a flat start of dynamic simulation, and at 1.0 s of the simulation time, a short-circuit fault is randomly applied at bus 4, 15 or 21 with a randomly-selected fault duration of 0.0 s (no fault), 0.05 s or 0.08 s; and the fault is self-cleared. This random selection of the fault location and duration could guarantee the training agent interacts with the system with and without FIDVR conditions. 
 \hl{The training process took 21 hours on the same Linux workstation used in the previous case without any paralization. }
 The moving average of the rewards   during the training is shown in Fig. \ref{fig:9}.
 
 After the training, we tested the robustness and adaptiveness of the trained DRL agent on a set of \hl{
960 test scenarios that have different combinations of power flow conditions, dynamic model parameters, fault locations, and fault duration from the training scenarios, as follows: (1) four different load levels (i.e., 80\%, 90\%,  110\%, and 120\% load levels); (2) two sets of critical dynamic parameters of the air-conditioner motor model, with one set corresponding to (assumed) true values and the other set considering a 10\% increase in the A/C motor stalling performance parameters $T_{stall}$ and $V_{stall}$ {\cite{Kosterev2008}}. Note that the air-conditioner motor dynamic model is an aggregated model that represents a large set of physical air-conditioners in the real environment,  so its parameters could contain many uncertainties; (3) 30 different fault locations (i.e., buses 1 to 30); and (4) four different fault duration times (i.e., 0.02, 0.05, 0.08 and 0.1 s). }
 
\hl{ We have compared the trained DRL-based load shedding control versus the UVLS relay load shedding scheme, as well as an MPC method that uses a mixed integer programming optimization to solve the problem described by ({\ref{eq: emergency_control_optimal}}).  We have compared all three control methods in terms of the execution time and the reward defined in ({\ref{eq: UVLS_reward}}). } To show the comparison results, we calculate the reward differences (i.e., the reward of DRL subtract that of a comparison method) for all the test scenarios, and a positive value means that the DRL method is better for the corresponding test scenario, and vice versa.

 \begin{figure}[!ht]
\centering
\subfigure[]{\includegraphics[width=0.45\textwidth]{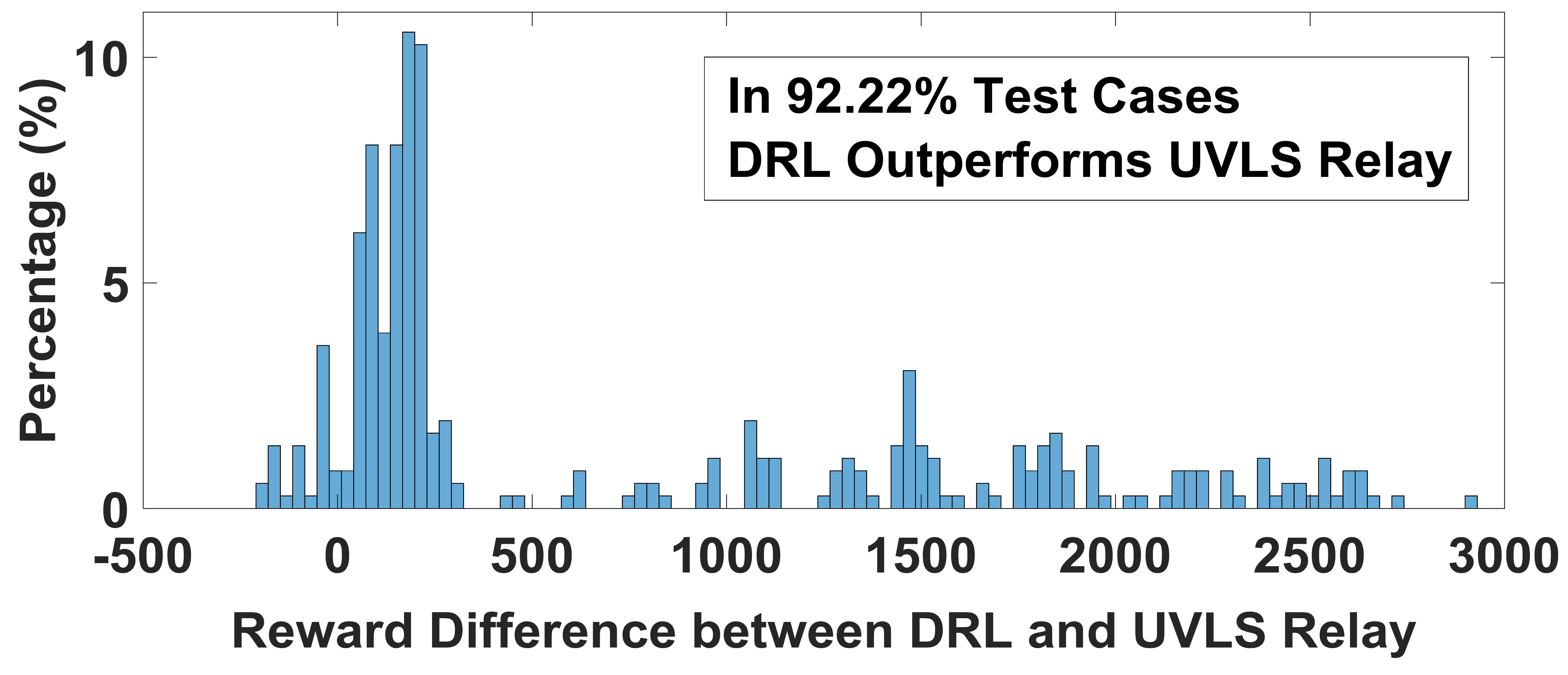}}
\subfigure[]{\includegraphics[width=0.45\textwidth]{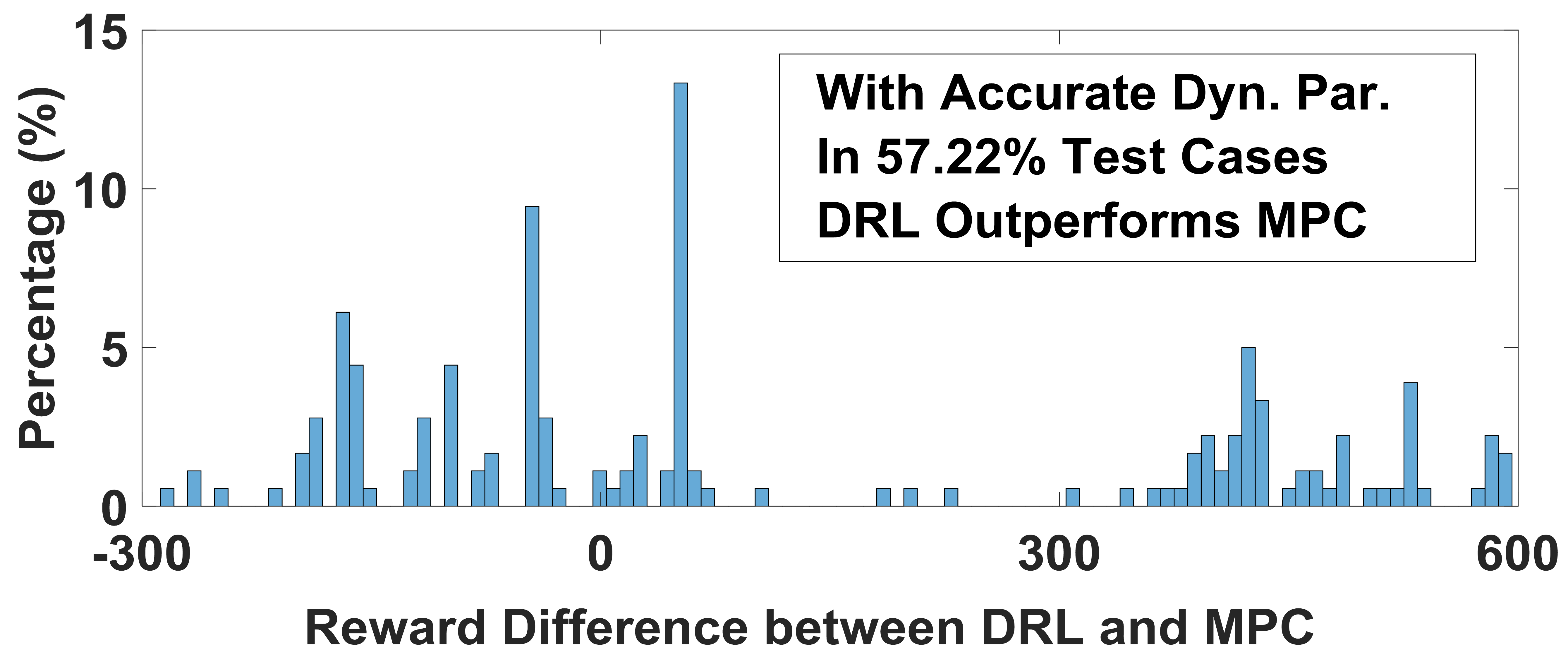}}
\subfigure[]{\includegraphics[width=0.45\textwidth]{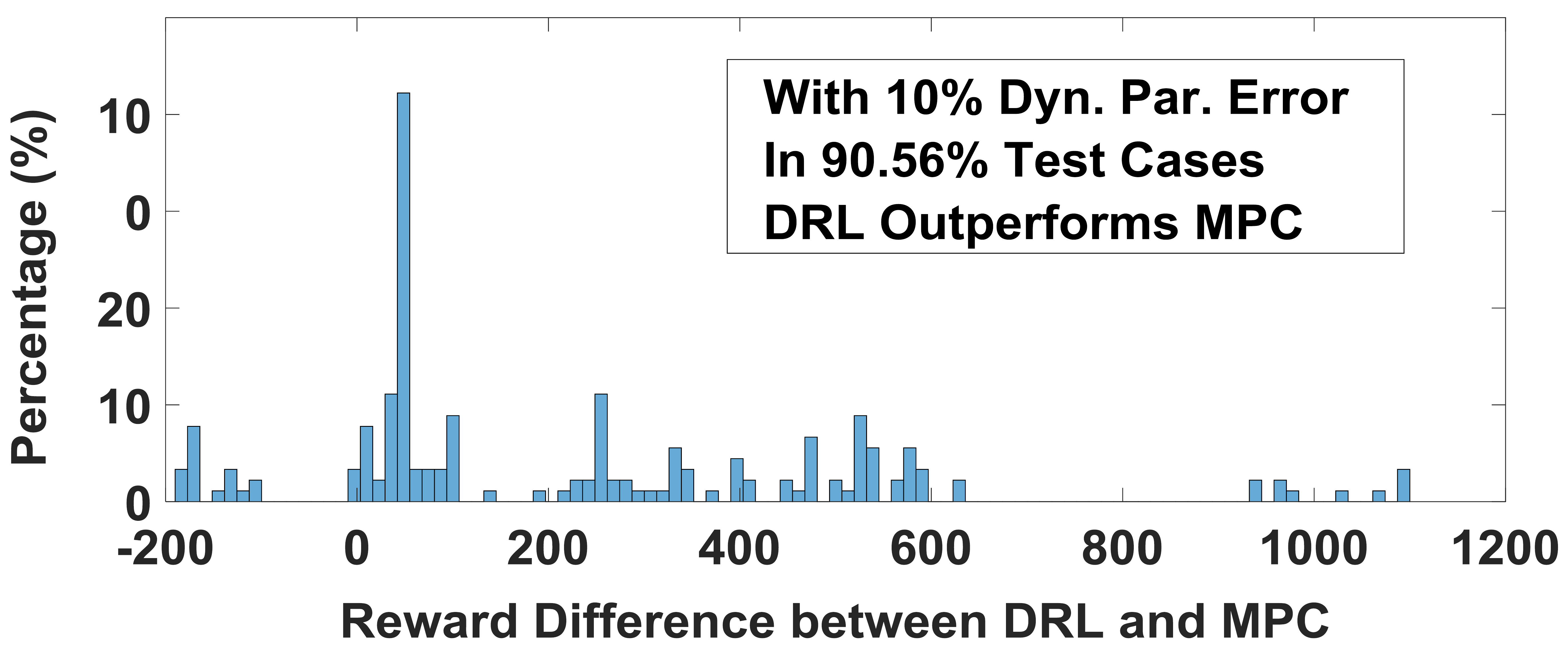}}
\caption{Histogram for reward difference between (a) DRL and UVLS for 462 test cases require load shedding; (b) DRL and MPC for 229 test cases in Test Set A; (c) DRL and MPC for 233 test cases in Test Set B }
\label{fig:histogram}
\vspace{-1mm} 
\end{figure}

 \hl{Among the 960 test scenarios, 462 of them could lead to FIDVR problems if no action is applied, and thus require load shedding. Fig. {\ref{fig:histogram}} (a) shows the histogram of the reward difference between the DRL-based control and the UVLS relay. The DRL-based control outperforms the UVLS relay for 92.22\% of these 462 test scenarios. Among the 462 test scenarios, 229 test scenarios have the same dynamic parameters as the training scenarios (Test Set A), while 233 test scenarios have a 10\% increase for the dynamic load parameters $T_{stall}$ and $V_{stall}$ (Test Set B). The main objective of Test Set B is to mimic the modeling gaps (or uncertainties) in real-world applications. Note that the DQN-based DRL method is model-free, while MPC-based methods heavily depend on the accuracy of the model; thus, it is important to consider the modeling errors in MPC-based applications.} 
 
  \hl{For Test Set A, Fig. {\ref{fig:histogram}} (b) depicts the histogram of reward difference between the DRL and MPC, which indicates that DRL-based control has a slightly better performance than the MPC (the DRL outperforms the MPC in 57.22\% of the test scenarios). For Test Set B, Fig. {\ref{fig:histogram}} (c) shows the histogram of the reward differences between the DRL and MPC methods, which  shows that the DRL method outperforms the MPC method in 90.56\% of the test scenarios.  Fig. {\ref{fig:histogram}} (b) and (c) clearly show a significant advantage of the developed DRL method over the MPC method: the performance of the MPC method heavily depends on the accuracy of the system model, while DRL is model-free and more robust to modeling errors.}
 
  \hl{Table {\ref{tab:table1}} shows the average computation time of the DRL and MPC methods. The computation time for UVLS relays is not included as it is either instantaneous or a predefined delay. It is clearly shown in Table {\ref{tab:table1}} that the DRL method requires much shorter execution time than the MPC method, because the NN handling the complex mapping from observed states to actions in the DRL approach is much more efficient compared to a time-consuming, complex optimization solution process in the MPC method. With 0.13 s action time during a 8-second simulation event, the DRL method can meet the real-time operation requirements and allows grid operators to verify the control actions when necessary. }
 
 \begin{table}[!ht]
 \renewcommand{\arraystretch}{1.2}
 \caption{Comparison of average computation time for the DRL and MPC}
     \centering
     \begin{tabular}{|c|c|}
     \hline  
       Average DRL Computation Time   &  Average MPC Computation Time  \\
       \hline  
        0.13 seconds  & 23.73 seconds \\
        \hline
     \end{tabular}
          \label{tab:table1}
 \end{table}
 
\begin{figure}[!ht]
\centering
\subfigure[]{\includegraphics[width=0.45\textwidth]{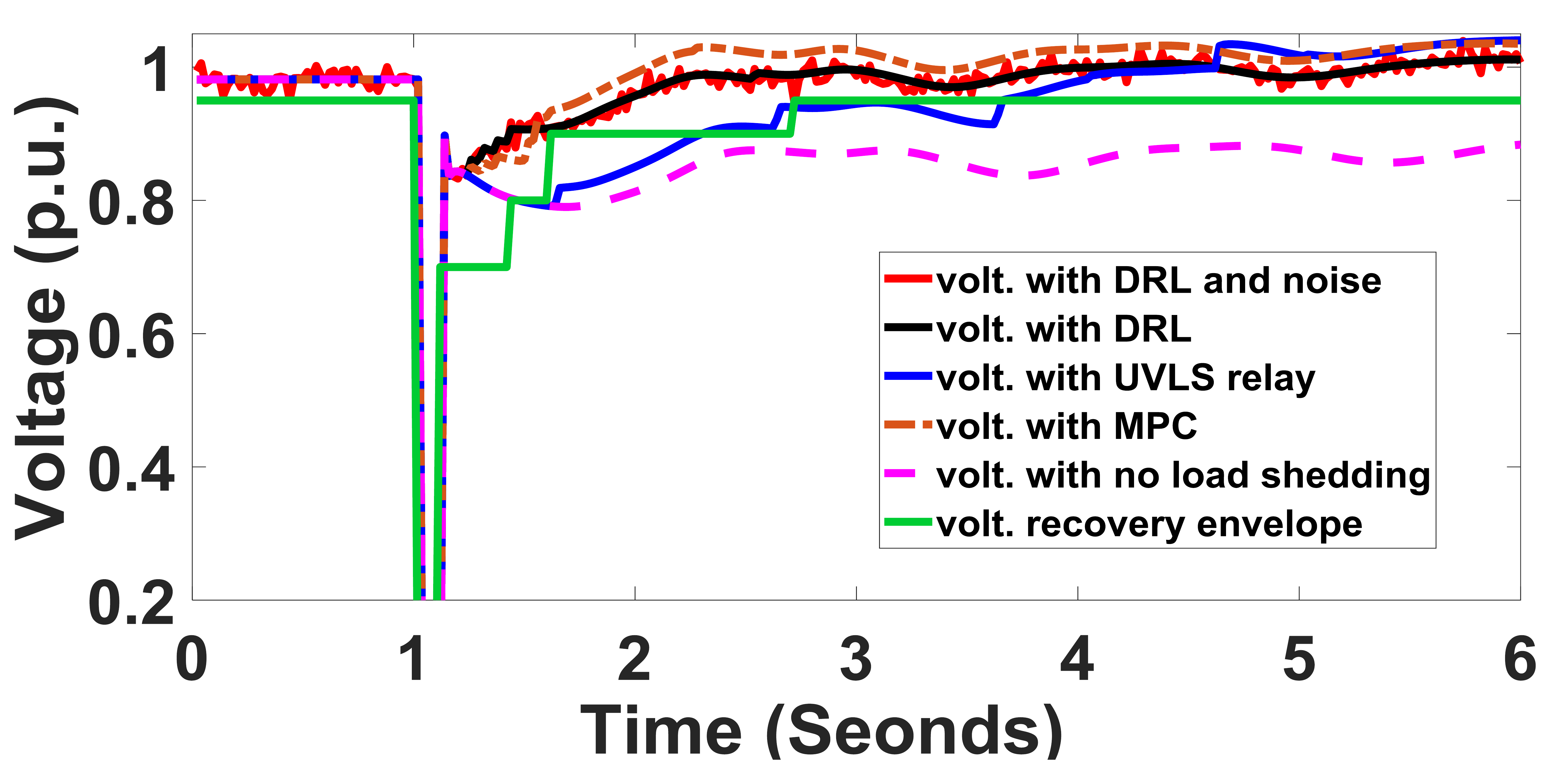}}
\subfigure[]{\includegraphics[width=0.45\textwidth]{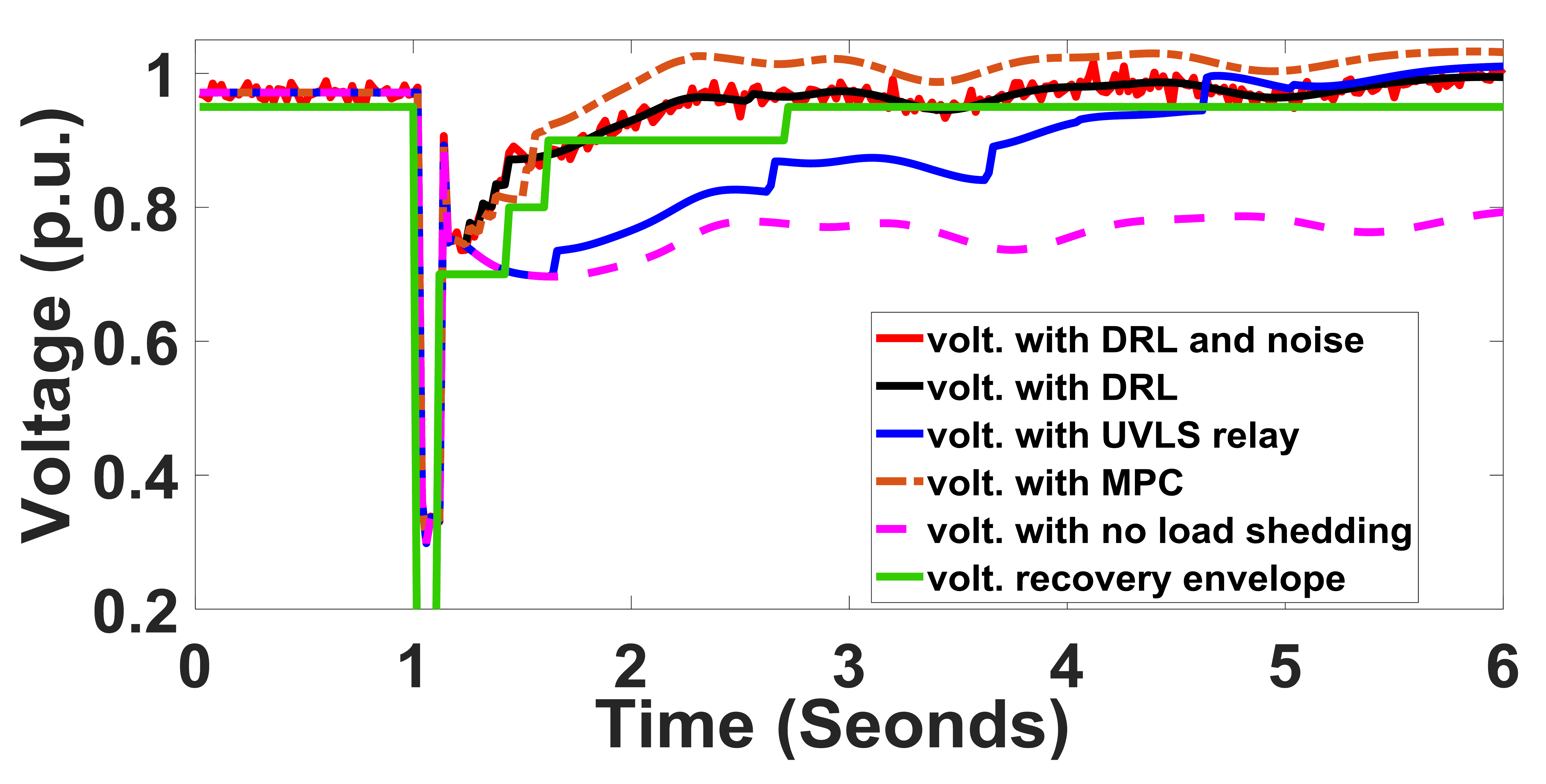}}
\subfigure[]{\includegraphics[width=0.45\textwidth]{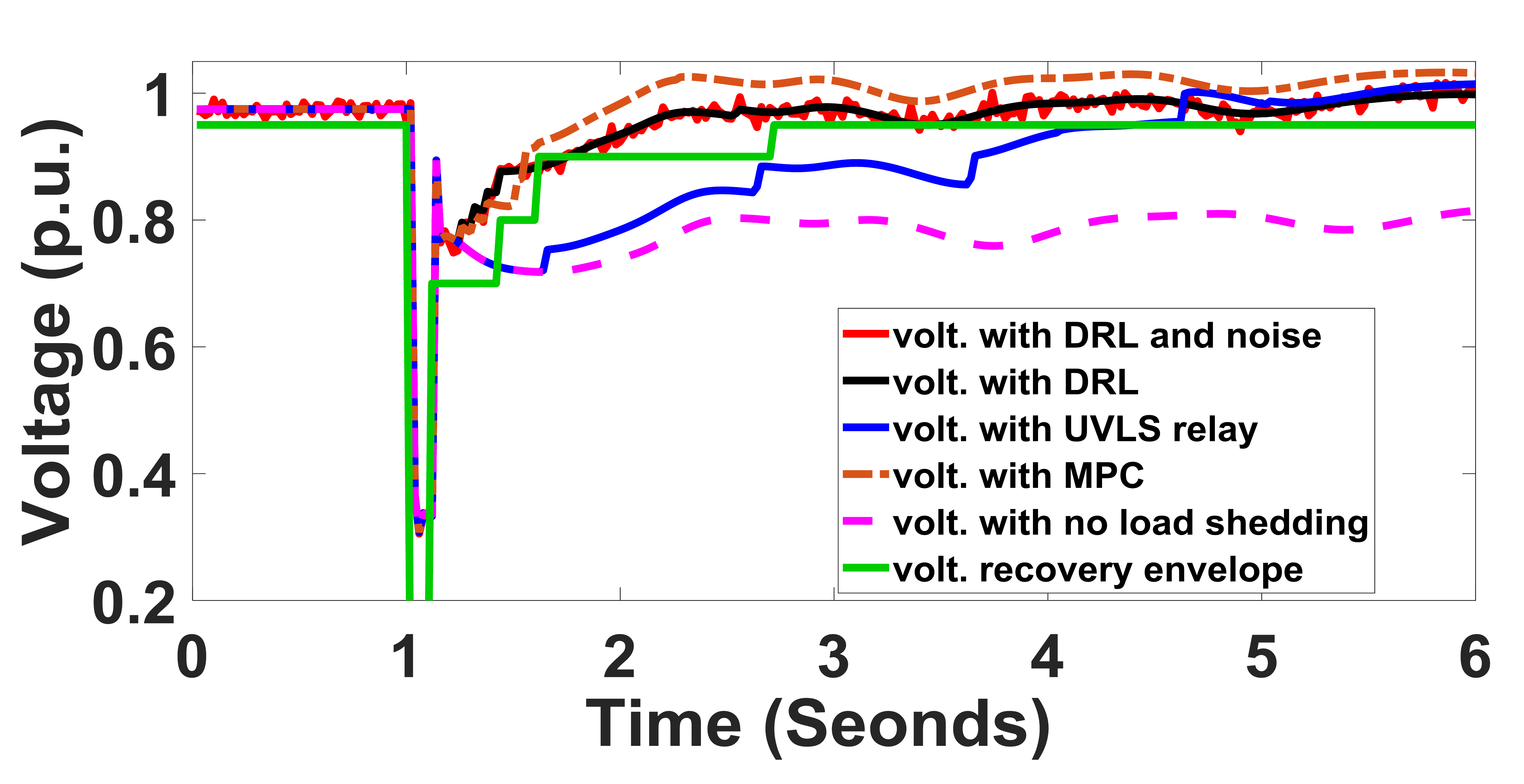}}
\caption{Voltage profiles for different load shedding control schemes and the required voltage recovery envelope: (a) bus 4; (b) bus 7; (c) bus 18}
\label{fig:10}
\vspace{-1mm} 
\end{figure}

\begin{figure}[!ht]
\centering
\subfigure[]{\includegraphics[width=0.45\textwidth]{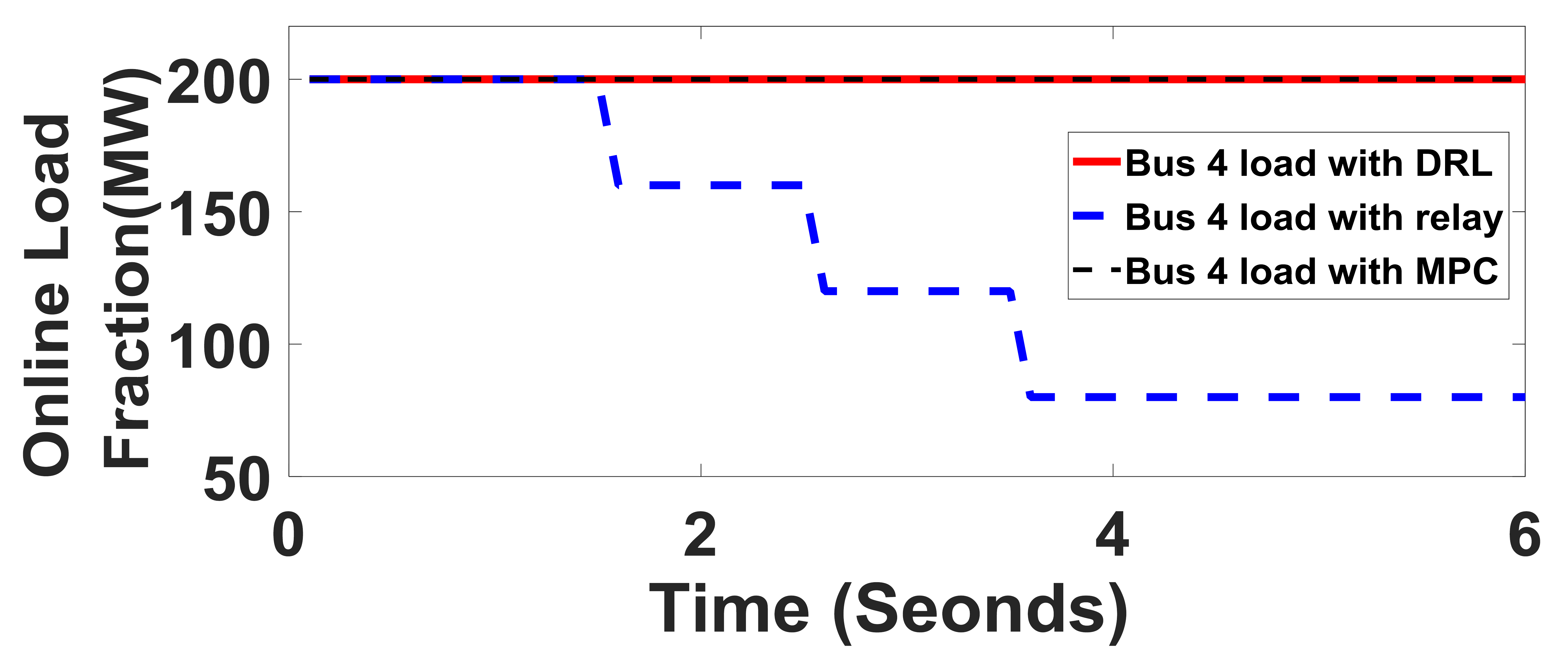}}
\subfigure[]{\includegraphics[width=0.45\textwidth]{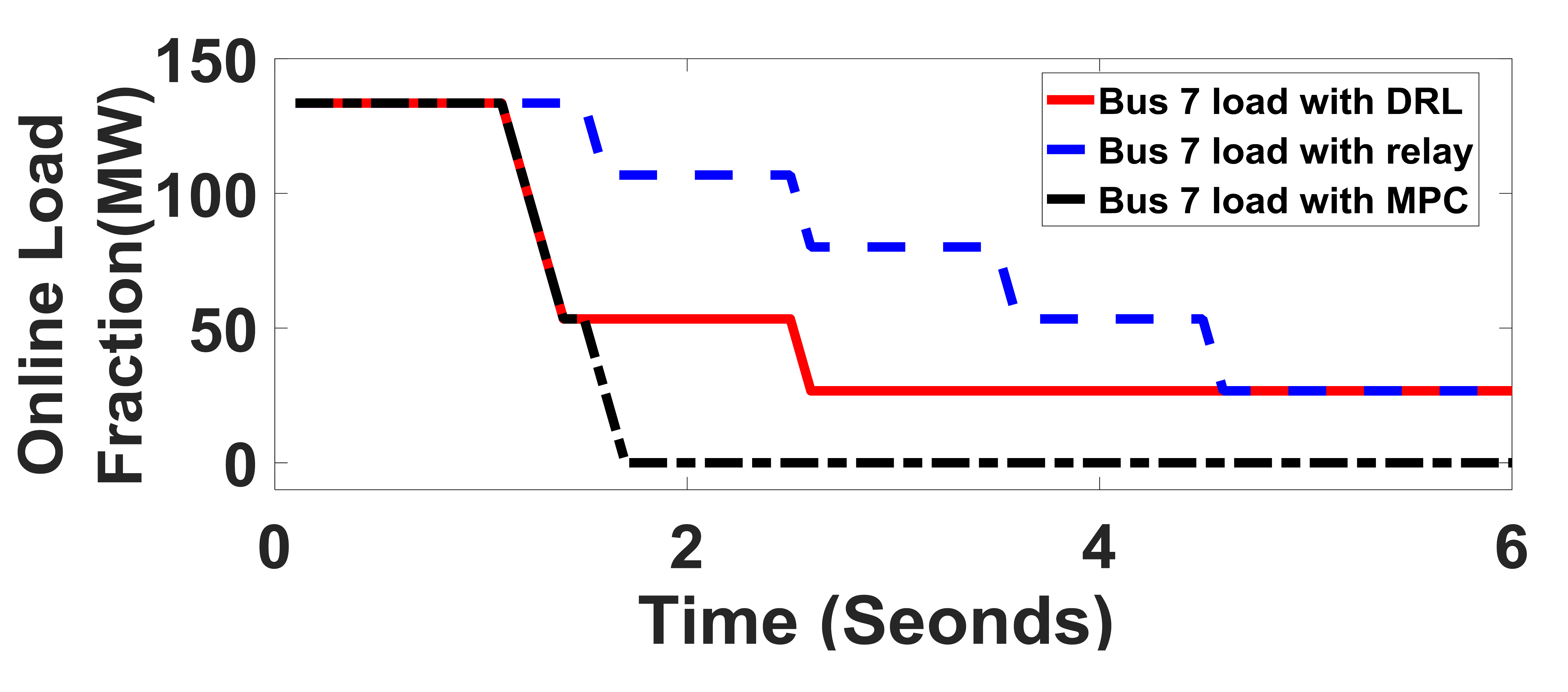}}
\subfigure[]{\includegraphics[width=0.45\textwidth]{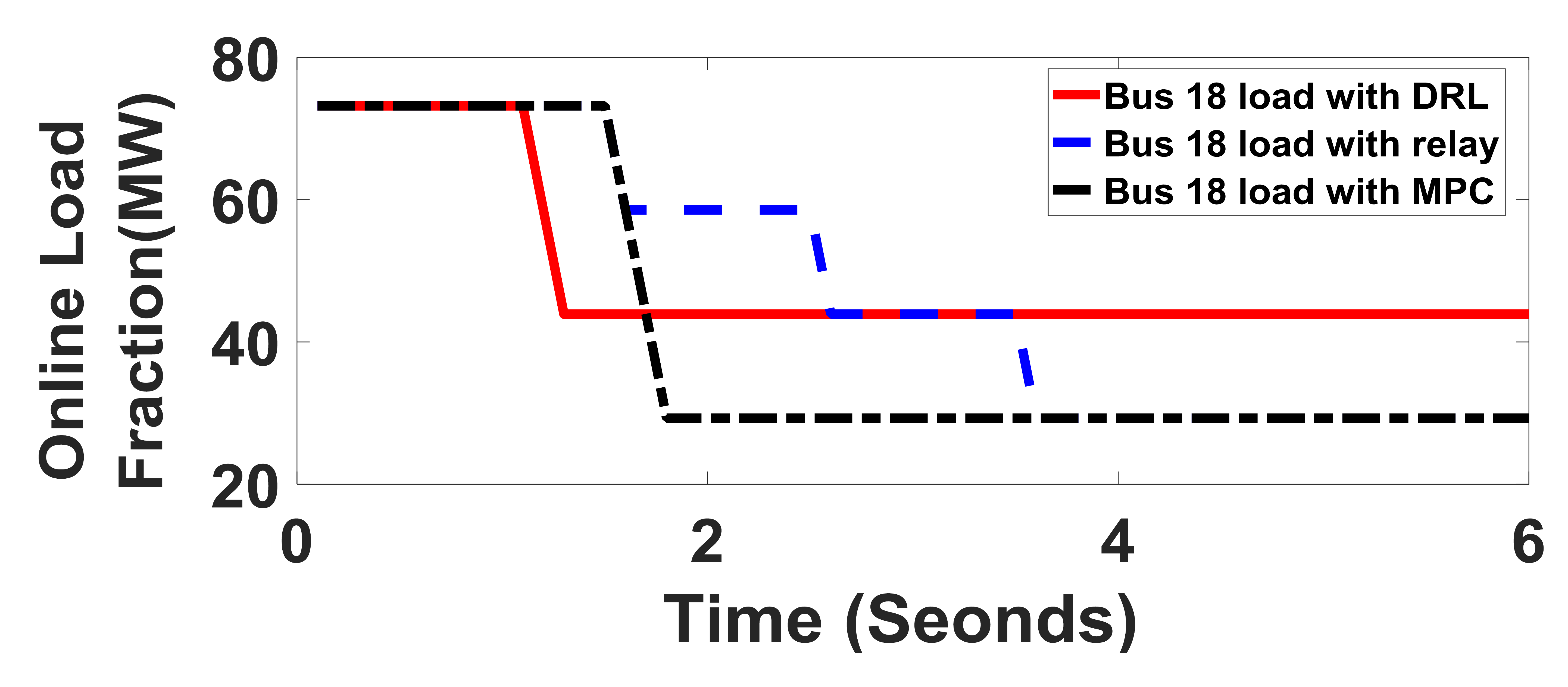}}
\caption{On-line load fractions with load shedding controlled by DRL, MPC, and UVLS relay schemes: (a) bus 4; (b) bus 7; (c) bus 18. The fractions are scaled by the initial bus load in MW.}
\label{fig:11}
\vspace{-1mm} 
\end{figure}

\hl{To further illustrate the advantages of the DRL method, Figs. {\ref{fig:10}} and {\ref{fig:11}} show the comparison of the performance of the DRL, MPC, and the UVLS relay control schemes for a new test scenario with 120\% load level. The fault occurs at bus 3 with a duration time of 0.1 s, and there is a 10\% increase in the dynamic parameters $T_{stall}$ and $V_{stall}$.  To make the testing for the DRL-based load shedding control more realistic, we also add zero mean, 1\% Gaussian-distributed noise to the observations. The total rewards of the DRL, MPC, and UVLS relay control in this test case are -1271.61, -1548.14, and -3778.80, respectively. Fig. {\ref{fig:10}} shows the voltage profiles at buses 4, 7, and 18 for different load shedding controls;  Fig. {\ref{fig:11}}   shows the load shedding amount at buses 4, 7, and 18 for the DRL, MPC, and UVLS relay control schemes. Note that the added 1\% noise does not affect the decision making and the performance of the DRL-based control. The large reward difference (2507.19) between the DRL and UVLS relay comes from two parts: 1) the DRL sheds a significantly less amount of loads than UVLS relay.  Fig. {\ref{fig:11}} shows that compared with the UVLS relay, the DRL sheds 60\% (120 MW) less load for bus 4 (the DRL method does not shed any load at bus 4) and 20\% (14.64 MW) less load for bus 18; 2) the DRL method leads to a much better voltage recovery profile compared with the UVLS relay method, as shown in Fig. {\ref{fig:10}}. With the DRL-based control, the voltages at all three load buses with the A/C motors recover quickly above the voltage recovery envelope required by the operation standard. In contrast, the UVLS relay method cannot recover the voltages at the three buses even at 3 s after the fault is cleared, which  causes the UVLS relays to shed more loads at these three buses. The reward difference (276.53) between the DRL and MPC methods is mainly due to the fact that the DRL method sheds less  load than the MPC while meeting the operation standard requirements. Fig. {\ref{fig:11}} shows that the DRL method sheds 20\% (26 MW) less load at bus 7, and 20\% (14.64 MW) less load at bus 18. The MPC method results in more load shedding as the MPC method suffers from inaccurate critical model parameters (10\% difference from the true values). Note that although Fig. {\ref{fig:10}}  shows that the voltage recovery profiles of the MPC method are slightly higher than the ones of the DRL method (at the cost of more loads being shed), this does not contribute to an increase of the reward, because the voltage profile being above the voltage recovery standard is not rewarded according to ({\ref{eq: UVLS_reward}}). We believe this is reasonable as the ultimate goal of UVLS controls is to recover the voltage above the envelope required by the standard with minimum load shedding.}

\hl{In summary, compared with the UVLS relay and MPC control methods, the DRL method shows significant improvements in terms of robustness and adaptiveness. In addition, the well-trained DRL model can provide control actions very fast (0.13 s on average) under emergency conditions, thus it can be applied for real-time emergency controls. }

\section{Discussions}\label{sec:discussion}

\hl{There are several important considerations for DRL application in general, and particularly in regards to its use in power system emergency control.}

\emph{1) Applicability to general emergency control problems:} We discussed how general power gridemergency control problems could be formulated as MDP problems and solved by DRL in Section II.C. Still, we believe that successful application of DRL to general emergency control problems heavily depends on properly formulating the problems as MDPs, including well-defined states, actions, and rewards. Given that automating the formulation process is still at an early research stage,   synergy between power domain knowledge and DRL, together with close collaborations between experts from both domains, is highly recommended.

\emph{2) Parameter selection:} In this paper, we manually tuned the parameters in the proposed algorithms, such as penalty factors and weighted factors in the reward functions. Determining these parameters is a known challenge for applying DRL and is also an active research topic in the RL community. 
Inspired by a recent work \cite{Chiang2019AutoRL}, we plan to automate this part in future efforts.

\emph{3) Reality gaps}: For controlling mission-critical infrastructures like power grids, 
training of the DRL agent(s) are, in general, performed in a simulation environment.
 There are always some reality gaps between models and real-world systems. One of the authors has made good progress in addressing this reality gap issue in the robotic domain {\cite{tan2018sim}}. We plan to adapt the developed technologies to solve power grid control problems in the future. 

\emph{4) Safety guarantee (or safe exploration)}: In this paper, operation and/or safety constraints are considered by adding appropriate violation penalties in the reward functions. Recently, constrained policy optimization {\cite{achiam2017constrained}}  and safe exploration {\cite{dalal2018safeRL}} methods were proposed to realize constrained reinforcement learning.

%%%%%%%%%%%%%%%%%%%%%%%%%%%%%%%%%%%%%%%%%%%%%%%%%%%%%%%%%%%%%%%%%%%%%%%%%%%%%%%%%%%%%%%%%%%%

\section{Conclusions and Future Work}\label{sec:Conc}
Emergency control is imperative to guarantee the secure and reliable operation of power systems, particularly under large disturbance or severe contingency conditions. This paper investigates developing adaptive emergency control schemes using DRL. To support the development and benchmarking of DRL algorithms for grid control, for the first time, an open-source platform named RLGC is developed. By open-sourcing it, we hope to provide a good starting point and an open benchmark that accelerates future research in this field. The platform is employed to develop two typical emergency control schemes, including dynamic generator brake and UVLS. The test results demonstrate the adaptiveness and robustness (to new scenarios, model parameter uncertainy and noise in observations) of the two developed DRL-based emergency control schemes, as well as the advantages over schemes based on conventional Q-learning, MPC and existing protection mechanisms.

Future research work includes: 1) functionality extension of the RLGC platform, for example, support of other power system simulators; 2) applying DRL for other emergency controls on larger-scale power systems and with continuous action spaces; 3) applying recent advancements such as safe exploration and deep meta-reinforcement learning to better address control challenges associated with increased uncertainties in power systems.

\section{Acknowledgement}\label{sec:acknoledgement}
The authors gratefully thank Dr. Guanji Hou for his valuable suggestions and assistance in developing the MPC-based emergency control method in this paper.

%\vspace{-4mm} 
\bibliographystyle{IEEEtran}
\bibliography{DRL}

\vfill

\end{document}